\documentclass[10pt]{article}
\usepackage{yalenlp}
\usepackage{fontawesome5}
\renewcommand{\lablogo}{}



\usepackage{wrapfig}
\usepackage{enumitem}
\usepackage{colortbl,tabularx}
\definecolor{hhTableInk}{HTML}{172720}
\definecolor{hhTableMuted}{HTML}{53625B}
\definecolor{hhTableRule}{HTML}{8D9E96}
\definecolor{hhTableHead}{HTML}{E4EFEB}
\definecolor{hhTableSubhead}{HTML}{F0F5F2}
\definecolor{hhTableGroup}{HTML}{E8EFF3}
\definecolor{hhTableStripe}{HTML}{F6F8F7}
\definecolor{hhTableAccent}{HTML}{E3F0E1}
\definecolor{hhTableAccentLight}{HTML}{EFF6EE}
\DeclareCaptionFont{hhTableInk}{\color{hhTableInk}}
\newcolumntype{H}{>{\raggedleft\arraybackslash}X}
\newcommand{\hhtablelayout}{%
  \color{hhTableInk}\arrayrulecolor{hhTableRule}%
  \setlength{\heavyrulewidth}{0.65pt}%
  \setlength{\lightrulewidth}{0.40pt}%
  \setlength{\cmidrulewidth}{0.35pt}%
  \setlength{\aboverulesep}{0.37ex}%
  \setlength{\belowrulesep}{0.35ex}%
  \renewcommand{\arraystretch}{1.12}%
}

\usepackage[framemethod=TikZ]{mdframed}
\usepackage{float,listings,textcomp}
\usepackage{pgfplots}
\pgfplotsset{compat=1.18}
\usepgfplotslibrary{groupplots}
\hypersetup{
  pdftitle={HyperHarness: Learning to Generate Personalized Agent Harnesses from User Histories},
  pdfauthor={Zhaoxuan Tan, Jiachen Zhao, Jingbo Yang, Zheyuan Liu, Chuhan Wang, Pei Zhou, Mengting Wan, Meng Jiang}
}

\newcommand{\ourmethod}{\textsc{HyperHarness}}

\title{\ourmethod{}: Learning to Generate\\
Personalized Agent Harnesses from User Histories}

\fancypagestyle{hyperharnessfirst}[yalefirst]{%
  \fancyhead[L]{%
    \raisebox{-11.5pt}[0pt][0pt]{%
      \includegraphics[height=21pt]{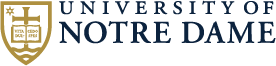}\hspace{12pt}%
      \raisebox{2pt}{\includegraphics[height=17pt]{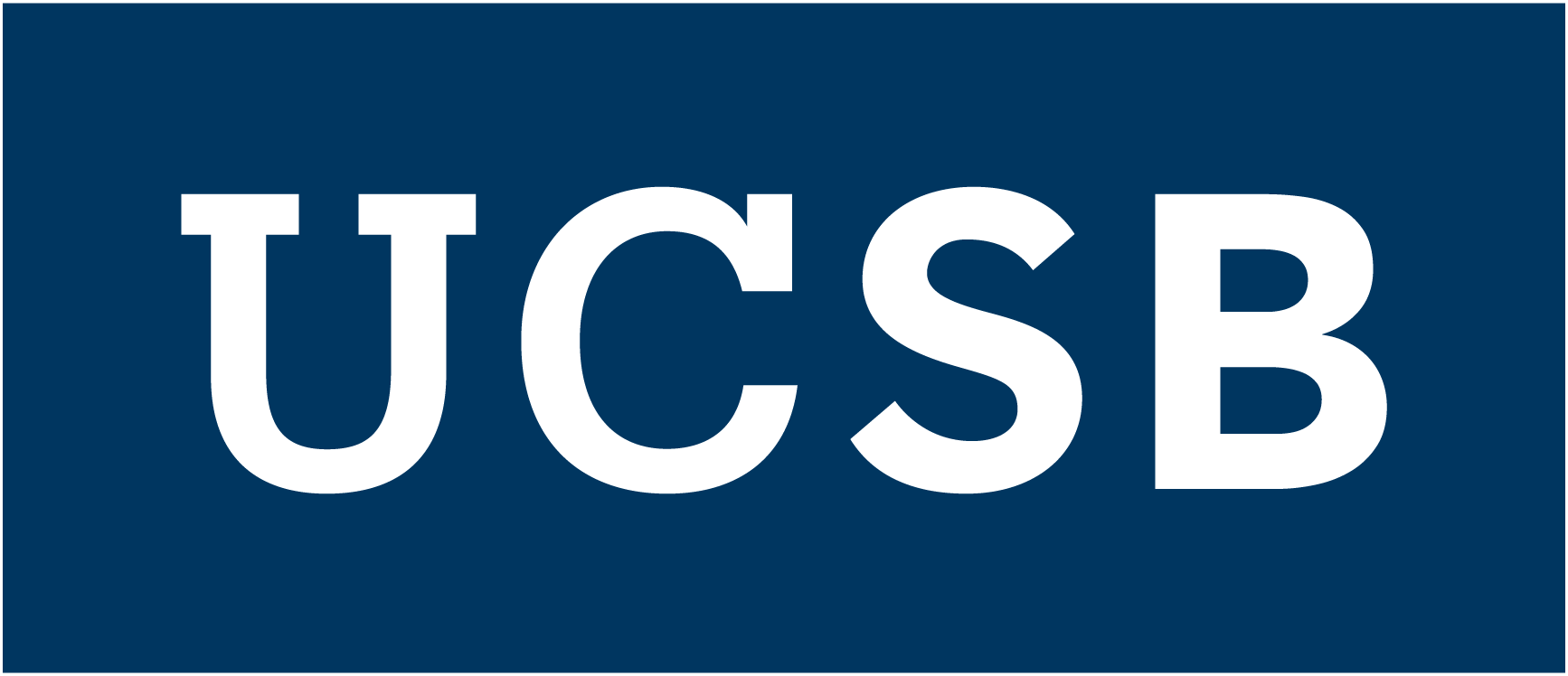}}\hspace{12pt}%
      \raisebox{1pt}{\includegraphics[height=19pt]{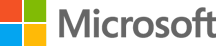}}%
    }%
  }%
}

\definecolor{hhAffiliation1}{HTML}{174E78}
\definecolor{hhAffiliation2}{HTML}{895513}
\definecolor{hhAffiliation3}{HTML}{684281}
\newcommand{\affiliationindex}[1]{\textsuperscript{\textcolor{hhAffiliation#1}{#1}}}

\author{%
  \textbf{Zhaoxuan Tan}\affiliationindex{1}\qquad
  \textbf{Jiachen Zhao}\affiliationindex{1}\qquad
  \textbf{Jingbo Yang}\affiliationindex{2}\qquad
  \textbf{Zheyuan Liu}\affiliationindex{1}\\[3pt]
  \textbf{Chuhan Wang}\affiliationindex{1}\qquad
  \textbf{Pei Zhou}\affiliationindex{3}\qquad
  \textbf{Mengting Wan}\affiliationindex{3}\qquad
  \textbf{Meng Jiang}\affiliationindex{1}\\[7pt]
  \textcolor{hhAffiliation1}{\textsuperscript{1}University of Notre Dame}\quad
  \textcolor{hhAffiliation2}{\textsuperscript{2}UC Santa Barbara}\quad
  \textcolor{hhAffiliation3}{\textsuperscript{3}Microsoft}\\[6pt]
  \href{mailto:ztan3@nd.edu}{\faEnvelope\enspace\texttt{ztan3@nd.edu}}\qquad
  \href{https://github.com/TamSiuhin/hyperharness-code}{\faGithub\enspace GitHub Code}
  \qquad
  \href{https://zhaoxuan.info/hyperharness/}{\faGlobe\enspace Project Website}
}
\date{}

\begin{document}
\maketitle
\thispagestyle{hyperharnessfirst}

\begin{abstract}
Personalized LLM agents rely on an \emph{agent harness}, the surrounding memory and control code, to connect user history with downstream decisions. Existing methods typically employ either a single shared memory workflow across all users, which is efficient to deploy but cannot adapt how evidence is retrieved and interpreted across diverse histories, or per-user harness optimization, which customizes behavior for each user but requires costly repeated candidate evaluation at deployment and suffers from sparse, noisy supervision in raw logs. To address this trade-off, we present \ourmethod{}, a framework that learns a shared history-to-harness generator to synthesize executable, user-specific retrieval and answer-guidance code directly from user histories. Given a user's permitted history, \ourmethod{} prompts a construction LLM to write personalized retrieval and guidance functions, while downstream task scores across training users supervise iterative refinements to the shared generator without reference harnesses or answering-model updates. At deployment, the frozen generator constructs a harness for each unseen user in a single pass from history alone, eliminating per-user task-scored search. Across three benchmarks (PersonaMemv3, Mem-PAL, and MemoryCD) and three answering models (\mbox{\texttt{Qwen3.5-9B}}, \mbox{\texttt{Qwen3.8-27B}}, and \mbox{\texttt{GPT-5.6-Luna}}), \ourmethod{} achieves the highest aggregate score in seven of nine settings and ranks second in the remaining two, improving over the strongest baselines by up to 5.95 points while cutting matched answer-time tokens by 90.32\% over long-context prompting. Analyses show that optimization tailors context allocation to each executor, transfers to unseen user clusters, and amortizes adaptation costs.
\end{abstract}

\section{Introduction}
\label{sec:introduction}
\begin{wrapfigure}{r}{0.55\textwidth}
  \vspace{-0.8\baselineskip}
  \centering
  \includegraphics[width=\linewidth]{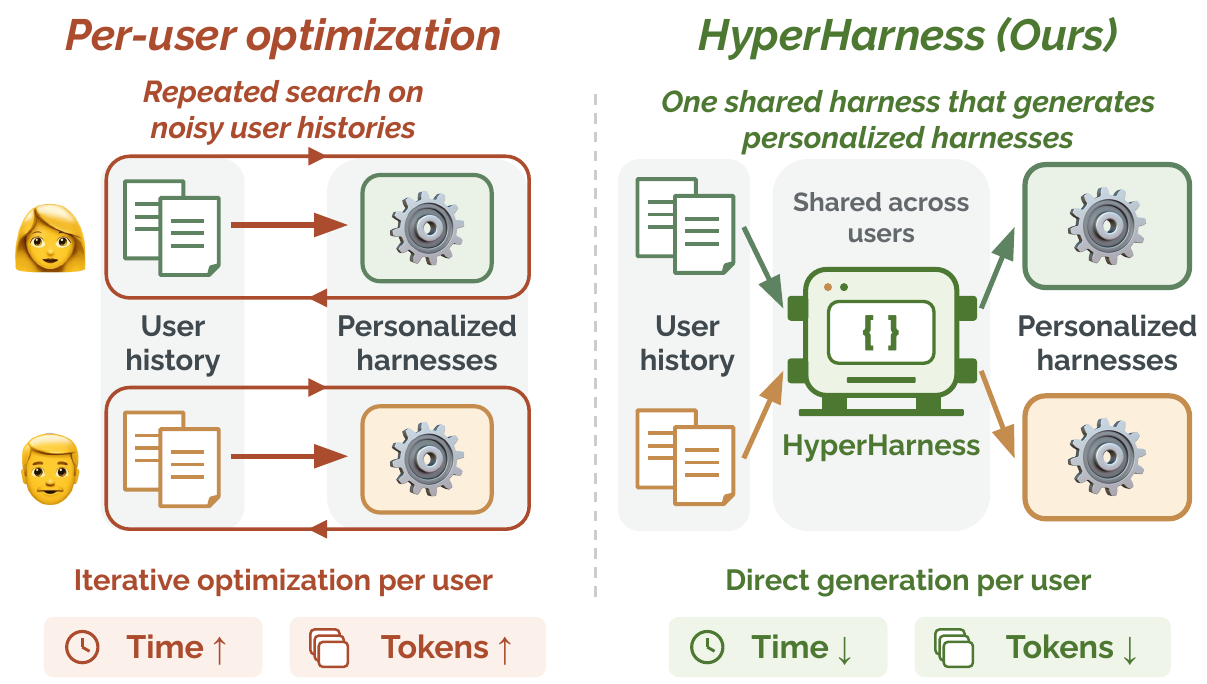}
  \caption{\ourmethod{} bridges shared workflows and per-user search with a history-to-harness generator for zero-search deployment.}
  \label{fig:teaser}
  \vspace{-0.6\baselineskip}
\end{wrapfigure}
Personalization is a core component of user-facing AI systems because it connects raw user behavior with downstream decisions about what content, recommendations, or responses should be presented to each individual \citep{tan2023usermodeling,kirk2024personalizing,salemi2024lamp,tan2024oppu}. As large language models (LLMs) evolve from static text generators into interactive agents, adapting to a user involves not only what the model says, but also the execution procedure used to reach an answer: which historical records to consult, how to reconcile conflicting or noisy evidence, and when to restrain personal context \citep{chen2024personalization,cai2025personalwab,yin2026usertoolbench}. In agentic systems, this decision process is governed by an \emph{agent harness}, the surrounding executable code and instructions that control how memory is stored, retrieved, and applied \citep{packer2023memgpt,xu2025amem}.

A central design choice in agent personalization is how to construct the harness that connects a user's interaction history to runtime execution. Existing approaches mainly follow two paradigms: shared memory workflows and per-user harness optimization. In practice, shared workflows, such as long-context prompting, fixed top-$k$ retrieval, or uniform rolling summaries \citep{salemi2024lamp,salemi2024retrieval,richardson2023summarization}, are the dominant choice because they can be deployed immediately to any user without per-user search. However, applying a single fixed procedure to every user overlooks substantial heterogeneity in how users express preferences and how their histories are structured. A compact summary may suffice for stable tastes, whereas shifting or multi-faceted interests require targeted multi-query retrieval and explicit disambiguation; simply packing more raw history into the prompt is token-intensive and frequently distracts the answering model with irrelevant records \citep{shi2023distracted}. In contrast, automated agent and harness optimization methods \citep{hu2025adas,zhang2025aflow,robeyns2025sica,zhang2025dgm,agrawal2026gepa,lee2026metaharness,yang2026scholarevolve} offer the flexibility to customize prompts or executable workflows through evaluation feedback. However, running task-scored search separately for each user faces two key challenges. First, repeated candidate generation and evaluation incur heavy token and latency overhead before the agent can serve a new user, often exceeding context budgets as histories grow. More importantly, an individual user's raw interaction log provides sparse and indirect supervision by mixing enduring preferences, temporary intent, and incidental exposure \citep{tan2025pugc,jiang2025personamemv2,tan2025perrecbench}, and there is no ground-truth reference harness specifying how that user's evidence should be retrieved or applied.

This exposes a clear trade-off between deployment efficiency and user-specific procedural adaptation: shared workflows are lightweight to deploy but rigid across users, whereas per-user search is expressive but costly and difficult to supervise at deployment. We resolve this trade-off by shifting optimization from individual users' harnesses to a shared construction procedure. We propose \ourmethod{}, a framework that learns a reusable history-to-harness generator capable of synthesizing executable, user-specific retrieval and answer-guidance code directly from a user's history (Figure~\ref{fig:teaser}). Inspired by hypernetworks that generate target network weights \citep{ha2017hypernetworks} and generation-based parameter adaptation \citep{phang2023hypertuning,ivison2023hint,tan2026profiletopeft}, \ourmethod{} extends amortized generation from static profiles or adapter weights to executable procedures that govern both which personal evidence is recalled and how it guides the answering model (Appendix~\ref{app:hypernetwork_discussion}).

Specifically, given a user's permitted construction history, \ourmethod{} organizes the historical records into structured evidence and prompts a frozen construction LLM to write user-specific code for memory retrieval and answer guidance under bounded host validity checks. During training, we keep both the construction and answering models frozen and optimize the shared generator's code and prompts across training users using downstream task scores. Much as textual user profiles can be optimized through downstream utility without ground-truth summaries \citep{gao2025langptune,tan2026blue}, supervising the generator via downstream task performance eliminates the need for reference harnesses while allowing the generator to discover which historical patterns call for which retrieval and guidance designs. At deployment, the selected generator is frozen and constructs a personalized harness for an unseen user in a single pass from history alone, requiring no task-scored per-user search and amortizing optimization cost across the user population.

We evaluate \ourmethod{} on PersonaMemv3 \citep{jiang2026personamemv3}, Mem-PAL \citep{huang2026mempal}, and MemoryCD \citep{zhang2026memorycd} across three frozen answering models (\mbox{\texttt{Qwen3.5-9B}}, \mbox{\texttt{Qwen3.8-27B}}, and \mbox{\texttt{GPT-5.6-Luna}}). Our results show that \ourmethod{} achieves the highest reported aggregate score in seven of nine configurations and ranks second in the remaining two, outperforming strong history, persistent memory, tool-using agent, and prompt/harness optimization baselines. With \mbox{\texttt{Qwen3.5-9B}}, \ourmethod{} improves over the strongest baseline by 5.95 points on PersonaMemv3 (54.51 versus 48.56) and 3.39 points on Mem-PAL (32.77 versus 29.38). In a matched execution comparison on MemoryCD, generated harnesses reduce mean answer-time tokens by 90.32\% relative to \textsc{Long Context} (3,147.28 versus 32,512.22 tokens per query) while preserving task performance, and amortize shared training costs to cross below per-user optimization baselines (\textsc{GEPA} and \textsc{Meta-Harness}) after a few dozen users. Extensive analyses further show that cross-user optimization adapts retrieval budgets and guidance structure to each answering model (e.g., gaining 7.96 points on PersonaMemv3 with \mbox{\texttt{Qwen3.5-9B}} while using 17.4\% fewer input tokens), transfers to unseen clusters on PersonaMemv3 and Mem-PAL, and remains competitive under sparse or noisy histories.

To summarize, our contributions are twofold. First, we propose \ourmethod{}, a cross-user learning framework that trains a shared history-to-harness generator from downstream task feedback, showing that user-specific retrieval and answer-guidance code can be synthesized directly from history at deployment without ground-truth reference harnesses, model weight updates, or per-user task-scored search. Second, through extensive experiments across assistant, preference-selection, and cross-domain recommendation benchmarks, we demonstrate that \ourmethod{} achieves strong personalization quality and a favorable efficiency frontier, while analyzing how learned harnesses adapt context allocation, generalize across user clusters, and respond to history perturbations.

\section{Methodology}
\label{sec:method}
\begin{figure}[!t]
\centering
\includegraphics[width=0.9\textwidth]{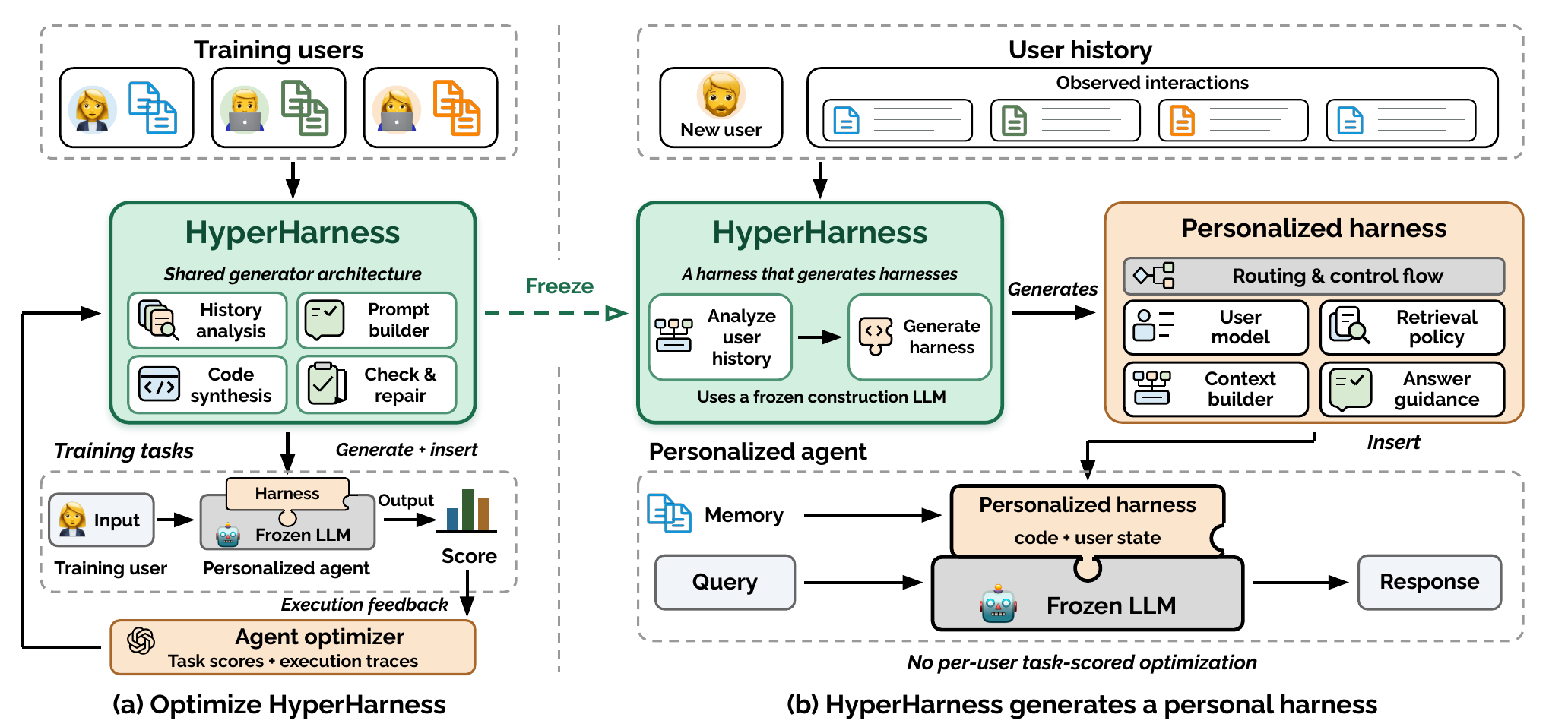}
\caption{Overview of \ourmethod{}. Given a user's permitted history, a shared generator organizes historical evidence and prompts a frozen construction LLM to synthesize executable retrieval and answer-guidance code. Downstream task scores across training users guide iterative updates to the generator specification, after which the frozen generator constructs personalized harnesses for unseen users directly from their histories without per-user search.}
\label{fig:overview}
\end{figure}

We aim to bridge rigid shared memory workflows and costly per-user harness search by learning a reusable construction procedure across users. Instead of treating harness generation as an isolated optimization problem solved separately for each user at deployment, we model it as a shared history-to-code generator that maps a user's behavioral history to executable retrieval and answer-guidance functions. The key idea is to use the downstream task utility of the generated harnesses as supervision: rather than matching ground-truth reference programs, the shared generator is optimized so that the harnesses it constructs improve downstream answers with a frozen answering model. In this pipeline, a user's permitted interaction history is first organized into structured evidence and converted into user-specific code, which is then executed to select and apply personal context at inference time, while the resulting task scores across training users are fed back to refine the generator specification (Figure~\ref{fig:overview}).

\subsection{Problem Formulation}
\label{sec:formulation}
Let $\mathcal U$ denote the set of users and $f_\theta$ denote a frozen answering model. For a given user $u\in\mathcal U$, let $\mathcal H_u^{\mathrm{con}}$ denote the interaction history permitted for harness construction. Our goal is to learn a shared generator $G_S$, parameterized by discrete code, prompts, and supporting resources $S$, that maps construction history to a personalized harness:
\begin{equation}
 h_u\sim G_S(\mathcal H_u^{\mathrm{con}};\mathcal C),
 \label{eq:harness-generation}
\end{equation}
where $\mathcal C$ specifies the runtime interface and execution budgets. Given a downstream request $x_{u,t}$ at time $t$, the personalized agent produces an answer by executing $h_u$ against the permitted memory store:
\begin{equation}
 \hat y_{u,t}=\operatorname{Run}(f_\theta,h_u,x_{u,t},z_{u,t}),\qquad
 z_{u,t}=\operatorname{Store}(\mathcal H_{u,t}^{\mathrm{avail}}).
 \label{eq:agent-execution}
\end{equation}
Here, the generated harness $h_u$ serves two complementary roles at runtime: (1) retrieving and organizing task-relevant evidence from the available history store $z_{u,t}$, and (2) generating user-specific guidance that conditions the frozen answering model $f_\theta$. The store $z_{u,t}$ contains only historical observations permitted by the benchmark protocol at time $t$; while $z_{u,t}$ may incorporate later chronological observations when allowed by the task, the constructed harness code $h_u$ remains fixed across all evaluation requests for user $u$. Training and test users are strictly disjoint, and construction never accesses test requests, reference answers, or test scores.

\subsection{Harness Generation and Execution}
\label{sec:architecture}
Figure~\ref{fig:overview} illustrates the overall generation and execution pipeline of \ourmethod{}. Starting from a user's permitted construction history $\mathcal H_u^{\mathrm{con}}$, the generator $G_S$ first organizes raw behavioral logs into structured evidence for a frozen construction LLM $g_\phi$. Whereas the initial generator specification uses a deterministic digest or bounded history preview paired with a shared runtime template, cross-user optimization can revise this history-processing code, the construction prompt, and the sequence of available search or read operations. Conditioned on the processed historical evidence and the interface specification $\mathcal C$, the construction model $g_\phi$ writes user-specific personalization code. Before a generated program is accepted, host-enforced interface and execution checks test its syntactic validity and runtime compliance using synthetic requests and permitted history, allowing bounded repair when execution errors arise. Crucially, these checks receive no personalization quality scores, cleanly separating executable validity from downstream task utility.

Once constructed, the generated harness $h_u$ functions as an executable bridge between the user's raw interaction history and the frozen answering model $f_\theta$. For each executor invocation with query $q$ and memory store $z$, the harness decomposes context construction into retrieval and answer guidance:
\begin{equation}
 c=h_u^{\mathrm{recall}}(q,z),\qquad
 a=h_u^{\mathrm{apply}}(q,c),\qquad
 \widetilde y\sim f_\theta(\cdot\mid q,c,a).
 \label{eq:personalized-context}
\end{equation}
Here, $c$ is the selected historical evidence and $a$ is the answer guidance specifying how to interpret and apply $c$ (with $a$ empty in retrieval-only settings); a downstream task may invoke this interface once or multiple times under its fixed protocol. To capture both targeted behavioral evidence and higher-level preference structure, $h_u^{\mathrm{recall}}$ can formulate custom search queries, filter noisy records, weight recency, and allocate the context budget, while $h_u^{\mathrm{apply}}$ can distinguish explicit preferences from weaker behavioral signals, reconcile conflicting evidence, or specify task-dependent priorities. Throughout execution, the host environment enforces fixed operation sets, context limits, and output validation, without introducing intermediate decision-model calls or post-hoc answer rewriting.

\paragraph{Editing Boundaries and Runtime Interface.}
To keep optimization modular and safe, \ourmethod{} enforces a strict three-layer separation between the host, the shared generator, and each user's personalized harness. First, the \emph{host runtime} is fixed and non-editable: it owns data boundaries, official scoring, execution budgets, and a restricted Python AST sandbox (plain functions without imports, classes, while loops, I/O, or intermediate model calls), interacting with construction through bounded propose-and-check turns and invoking harnesses solely via the top-level \texttt{recall} and \texttt{apply} entry points. Second, the \emph{shared generator} $G_S$ is the only layer edited by cross-user optimization; it bundles history digestion code, the construction prompt and call budgets, and a shared runtime backbone that implements common retrieval, packing, and dispatch logic around a single personal-section insertion slot. Third, once $G_S$ is frozen, that shared backbone becomes read-only to the per-user constructor, and host AST validation restricts each \emph{personalized harness} $h_u$ to defining a user-state dictionary (\texttt{USER\_MODEL}) and standard hook functions: \texttt{user\_route} (retrieval triggers, queries, and per-request depth), \texttt{user\_keep} (record filtering), \texttt{user\_guidance} (answer conditioning), and optional dispatch-gated overrides (\texttt{u\_recall}, \texttt{u\_apply}). Personal sections cannot redefine shared entry points, read shared global state, or relax host constraints, cleanly separating cross-user design edits from per-user specialization (full templates, neutral hook signatures, and generated code examples appear in Appendix~\ref{app:prompts_and_harness}).

\subsection{Training and Inference}
\label{sec:training-inference}
\paragraph{Cross-User Generator Optimization.}
Unlike supervised code generation, personalizing an agent harness admits no single ground-truth program for each user, as different combinations of retrieval logic and guidance can effectively support a user's preferences. We therefore optimize the shared generator specification $S$ based on functional downstream quality. Let $m_u(h)$ denote the downstream task score of harness $h$ on training user $u\in\mathcal U_{\mathrm{tr}}$. We seek a generator specification that maximizes the expected equal-user mean performance across training users:
\begin{equation}
 S^*\in\arg\max_S J(S),\qquad 
 J(S)=\frac{1}{|\mathcal U_{\mathrm{tr}}|}\sum_{u\in\mathcal U_{\mathrm{tr}}} 
 \mathbb E_{h\sim G_S(\mathcal H_u^{\mathrm{con}};\mathcal C)}[m_u(h)]. 
 \label{eq:generator-objective} 
\end{equation}
This objective supervises the generator directly through downstream task feedback without requiring reference harnesses or updating the weights of $g_\phi$ or $f_\theta$. Because $S$ comprises discrete code and prompts, we optimize Equation~\eqref{eq:generator-objective} via an iterative LLM-based search. Starting from an initial specification, the optimizer inspects the current generator, the synthesized user programs, their downstream scores, and diagnostic execution traces, and proposes a revised construction procedure. At round $r$, each candidate specification constructs fresh harnesses for the training cohort and is compared against the current incumbent on the same evaluation panel $\mathcal P_r$:
\begin{equation}
 \widehat J_r(S)=\frac{1}{|\mathcal P_r|}\sum_{u\in\mathcal P_r}m_u(h_{u,S}), 
 \qquad h_{u,S}\sim G_S(\mathcal H_u^{\mathrm{con}};\mathcal C). 
 \label{eq:paired-selection} 
\end{equation}
An eligible candidate replaces the incumbent when it achieves a higher panel mean score, while ties retain the incumbent (benchmark-specific selection criteria are detailed in Appendix~\ref{app:implementation}). Within fixed host boundaries, the optimizer can jointly refine history preprocessing, construction prompts, and runtime control flow.

\paragraph{Zero-Search Inference at Deployment.}
Once the training search budget is exhausted, we freeze the selected generator specification $\widehat S$. At inference time, serving an unseen test user $v\notin\mathcal U_{\mathrm{tr}}$ proceeds in two stages without any task-scored search:
\begin{itemize}[leftmargin=*]
\item \textbf{One-time per-user harness construction:} Given user $v$'s permitted construction history $\mathcal H_v^{\mathrm{con}}$, the frozen generator synthesizes a user-specific harness $h_v\sim G_{\widehat S}(\mathcal H_v^{\mathrm{con}};\mathcal C)$ in a single construction pass. Host-enforced syntax and execution checks ensure a valid program, while no test queries, reference answers, or scores from user $v$ are accessed.
\item \textbf{Online query execution:} The constructed harness $h_v$ is then held fixed across all incoming test requests $\{x_{v,t}\}$ for user $v$. For each request $x_{v,t}$, the runtime updates the permitted memory store $z_{v,t}=\operatorname{Store}(\mathcal H_{v,t}^{\mathrm{avail}})$, executes $c_{v,t}=h_v^{\mathrm{recall}}(q_{v,t},z_{v,t})$ and $a_{v,t}=h_v^{\mathrm{apply}}(q_{v,t},c_{v,t})$ to assemble the personalized context, and queries the frozen answering model $\widetilde y_{v,t}\sim f_\theta(\cdot\mid q_{v,t},c_{v,t},a_{v,t})$.
\end{itemize}
By decoupling cross-user training, one-time harness construction, and lightweight query execution, \ourmethod{} amortizes optimization overhead across users and eliminates per-user search at deployment.

\section{Experimental Settings}
\label{sec:experiments}
\label{sec:experimental_setup}
\subsection{Models and Datasets}
We instantiate \ourmethod{} using \texttt{gpt-6-sol} \citep{openai2026gpt6sol} for harness construction and generator optimization, and evaluate generated harnesses with three frozen answering models: \texttt{Qwen3.5-9B}, \texttt{Qwen3.8-27B}, and \texttt{GPT-5.6-Luna} \citep{qwen2026qwen35nine,qwen2026qwen38twentyseven,openai2026gpt56}. We conduct experiments on three benchmarks covering complementary personalization tasks with strictly disjoint training and test users. \textbf{PersonaMemv3} \citep{jiang2026personamemv3} evaluates personalized assistant interactions across recommendations, evolving preferences, restraint, and proactive actions (25 training and 75 test users). \textbf{Mem-PAL} \citep{huang2026mempal} evaluates preference-based solution selection on 50 training and 50 test users (408 test questions). \textbf{MemoryCD} \citep{zhang2026memorycd} evaluates cross-domain transfer from Books, Electronics, and Beauty histories to held-out Home tasks spanning item ranking, rating prediction, and review generation (100 training and 223 test users; 2,676 test instances). During construction, the generator accesses only the permitted historical prefix or source-domain history and never observes test queries, reference answers, or test scores. Table~\ref{tab:data_statistics} summarizes the experimental splits.

\subsection{Baselines and Evaluation}
\paragraph{History and Memory Baselines.}
We first compare against direct history-access and persistent memory methods, with \textsc{No History} as a reference. \textsc{Long Context} packs the most recent history events within the input budget, and \textsc{BM25} retrieves the top ten relevant records per request \citep{robertson2009bm25}. For persistent memory, \textsc{Rolling Memory} maintains a summary updated every 20 events, and \textsc{Mem0 with RAG} constructs memories with the answering model and retrieves ten entries via \texttt{MiniLM} \citep{chhikara2025mem0,wang2020minilm}. These baselines test whether generated harnesses improve over shared retrieval and memory-compression workflows.

\paragraph{Agent and Optimization Baselines.}
We further compare against \textsc{Codex CLI}, which runs the answering model as a tool-using agent \citep{openai2026codexcli}, \textsc{GEPA} for reflective prompt optimization \citep{agrawal2026gepa}, and \textsc{Meta-Harness} for trace-guided executable harness optimization \citep{lee2026metaharness}. These methods test whether single-pass harness generation remains competitive with multi-step agent execution and task-scored search.

\paragraph{Evaluation.}
For each benchmark, we compute official task metrics per user and average equally across users: rule-based and \texttt{gpt-5.5} scores on PersonaMemv3, signed hit-rate score on Mem-PAL, and the equal average of ranking NDCG@5, normalized rating accuracy, and review ROUGE-L on MemoryCD \citep{jarvelin2002ndcg,lin2004rouge}. We also report deployment token usage $T_{\mathrm{deploy}}$ in thousands of tokens per scheduled query, combining per-user setup and online execution while excluding judging and shared generator training. Full metric definitions, execution budgets, and implementation details are provided in Appendix~\ref{app:implementation}.

\section{Results}
\label{sec:main_results}
Table~\ref{tab:results} compares \ourmethod{} against history, persistent memory, agent, and optimization baselines across three benchmarks and three answering models. \ourmethod{} achieves the highest aggregate score in seven of nine configurations and ranks second in the remaining two. We highlight three main observations below.

\begin{table}[!t]\centering
\caption{Main personalization results across three benchmarks and three frozen answering models. We report user-equal TEST scores ($\uparrow$) and deployment tokens $T_{\mathrm{deploy}}$ per scheduled query in thousands ($\downarrow$). Bold and underline mark the best and second-best values for each metric among personalized methods. Gray indicates the \textsc{No History} reference.}
\label{tab:results}
\begingroup
\footnotesize
\setlength{\tabcolsep}{4pt}
\renewcommand{\arraystretch}{1.12}
\hhtablelayout
\begin{tabularx}{\linewidth}{l*{6}{H}}
\toprule
\rowcolor{hhTableSubhead}& \multicolumn{2}{c}{\texttt{Qwen3.5-9B}} & \multicolumn{2}{c}{\texttt{Qwen3.8-27B}} & \multicolumn{2}{c}{\texttt{GPT-5.6-Luna}} \\
\cmidrule(lr){2-3}\cmidrule(lr){4-5}\cmidrule(lr){6-7}
\rowcolor{hhTableHead}\textbf{Method} & Score $\uparrow$ & $T_{\mathrm{deploy}}\downarrow$ & Score $\uparrow$ & $T_{\mathrm{deploy}}\downarrow$ & Score $\uparrow$ & $T_{\mathrm{deploy}}\downarrow$ \\
\midrule
\rowcolor{hhTableGroup}\multicolumn{7}{l}{\textbf{\textit{PersonaMemv3}}} \\
\rowcolor{hhTableStripe}\textcolor{hhTableMuted}{\textsc{No History}} & \textcolor{hhTableMuted}{45.54} & \textcolor{hhTableMuted}{0.79} & \textcolor{hhTableMuted}{47.46} & \textcolor{hhTableMuted}{0.84} & \textcolor{hhTableMuted}{53.90} & \textcolor{hhTableMuted}{0.69} \\
\rowcolor{white}\textsc{Long Context} & \underline{48.56} & 38.23 & 51.66 & 38.24 & 55.81 & 38.01 \\
\rowcolor{hhTableStripe}\textsc{BM25} & 46.79 & \underline{4.42} & 51.40 & \underline{4.47} & 56.00 & \underline{4.21} \\
\rowcolor{white}\textsc{Rolling Memory} & 44.51 & 23.97 & 49.03 & 24.12 & \underline{57.79} & 22.21 \\
\rowcolor{hhTableStripe}\textsc{Mem0 with RAG} & 44.42 & 25.83 & 48.40 & 26.38 & 56.09 & 26.90 \\
\rowcolor{white}\textsc{Codex CLI} & 40.41 & 40.13 & 47.45 & 63.37 & 55.14 & 31.46 \\
\rowcolor{hhTableStripe}\textsc{GEPA} & 47.93 & 38.24 & \textbf{53.14} & 38.30 & 56.47 & 37.95 \\
\rowcolor{white}\textsc{Meta-Harness} & 45.74 & 8.09 & 52.55 & 28.46 & 55.81 & 5.96 \\
\rowcolor{hhTableAccent}\ourmethod{} & \textbf{54.51} & \textbf{2.07} & \underline{52.83} & \textbf{4.17} & \textbf{58.59} & \textbf{2.23} \\
\midrule
\rowcolor{hhTableGroup}\multicolumn{7}{l}{\textbf{\textit{Mem-PAL}}} \\
\rowcolor{hhTableStripe}\textcolor{hhTableMuted}{\textsc{No History}} & \textcolor{hhTableMuted}{27.08} & \textcolor{hhTableMuted}{3.98} & \textcolor{hhTableMuted}{26.48} & \textcolor{hhTableMuted}{3.98} & \textcolor{hhTableMuted}{27.74} & \textcolor{hhTableMuted}{3.46} \\
\rowcolor{white}\textsc{Long Context} & 28.96 & 32.72 & 26.06 & 32.73 & 28.70 & 32.37 \\
\rowcolor{hhTableStripe}\textsc{BM25} & 27.43 & \textbf{5.02} & 27.64 & \textbf{5.03} & 27.85 & \textbf{4.40} \\
\rowcolor{white}\textsc{Rolling Memory} & 28.92 & 63.82 & 29.39 & 63.94 & \underline{30.92} & 54.76 \\
\rowcolor{hhTableStripe}\textsc{Mem0 with RAG} & 25.00 & 73.68 & 26.69 & 88.46 & 27.47 & 100.09 \\
\rowcolor{white}\textsc{Codex CLI} & 20.86 & 9.37 & 27.19 & \underline{9.36} & 29.22 & 11.12 \\
\rowcolor{hhTableStripe}\textsc{GEPA} & 26.50 & 32.79 & \textbf{30.08} & 32.73 & 29.77 & 32.41 \\
\rowcolor{white}\textsc{Meta-Harness} & \underline{29.38} & \underline{7.52} & 29.26 & 13.44 & 28.74 & 8.93 \\
\rowcolor{hhTableAccent}\ourmethod{} & \textbf{32.77} & 11.82 & \underline{30.04} & 11.58 & \textbf{31.27} & \underline{8.33} \\
\midrule
\rowcolor{hhTableGroup}\multicolumn{7}{l}{\textbf{\textit{MemoryCD}}} \\
\rowcolor{hhTableStripe}\textcolor{hhTableMuted}{\textsc{No History}} & \textcolor{hhTableMuted}{32.59} & \textcolor{hhTableMuted}{0.96} & \textcolor{hhTableMuted}{28.19} & \textcolor{hhTableMuted}{0.96} & \textcolor{hhTableMuted}{30.03} & \textcolor{hhTableMuted}{0.96} \\
\rowcolor{white}\textsc{Long Context} & \underline{34.94} & 32.51 & \underline{33.64} & 32.54 & 37.84 & 32.24 \\
\rowcolor{hhTableStripe}\textsc{BM25} & 33.09 & \textbf{2.54} & 31.05 & \textbf{2.55} & 36.65 & \textbf{2.45} \\
\rowcolor{white}\textsc{Rolling Memory} & 23.34 & 35.40 & 28.75 & 35.54 & 34.69 & 32.55 \\
\rowcolor{hhTableStripe}\textsc{Mem0 with RAG} & 32.81 & 37.14 & 28.30 & 40.09 & 34.16 & 37.78 \\
\rowcolor{white}\textsc{Codex CLI} & 27.95 & 6.35 & 27.01 & 6.35 & 31.60 & 17.49 \\
\rowcolor{hhTableStripe}\textsc{GEPA} & 34.24 & 32.51 & 32.22 & 32.54 & \underline{38.31} & 32.24 \\
\rowcolor{white}\textsc{Meta-Harness} & 33.29 & \underline{4.23} & 31.51 & 12.14 & 36.57 & \underline{4.00} \\
\rowcolor{hhTableAccent}\ourmethod{} & \textbf{35.23} & 4.96 & \textbf{34.51} & \underline{4.57} & \textbf{38.56} & 4.46 \\
\bottomrule
\end{tabularx}
\par\smallskip
\endgroup
\end{table}

\paragraph{Executable personalization outperforms shared history and memory workflows.}
Across benchmarks, generating user-specific retrieval and guidance code is consistently more effective than applying a uniform context workflow to every user. Packing raw history in \textsc{Long Context} exposes the answering model to irrelevant interactions that dilute task-critical signals, whereas fixed lexical retrieval in \textsc{BM25} struggles when preferences span multiple topics or require implicit reasoning. Meanwhile, persistent memory methods (\textsc{Rolling Memory} and \textsc{Mem0 with RAG}) perform well only when paired with a strong backbone capable of maintaining faithful summaries (such as \textsc{Rolling Memory} with \mbox{\texttt{GPT-5.6-Luna}}), but degrade sharply on smaller models where iterative compression discards fine-grained evidence. By synthesizing tailored retrieval and guidance logic for each user, \ourmethod{} adapts evidence selection to the structure of the user's history and achieves the best score across all three benchmarks with both \mbox{\texttt{Qwen3.5-9B}} and \mbox{\texttt{GPT-5.6-Luna}}.

\paragraph{Amortized generation matches or exceeds per-user search by regulating preference use.}
Despite constructing harnesses in a single pass without test-time search, \ourmethod{} outperforms the multi-step \textsc{Codex CLI} agent and matches or surpasses the reflective optimizers \textsc{GEPA} and \textsc{Meta-Harness}. Unconstrained agent exploration in \textsc{Codex CLI} often struggles to extract stable preference signals from noisy logs even with multiple calls per query. In contrast, cross-user generator training learns a strong structural prior that generalizes directly to unseen users, yielding particularly clear gains for \mbox{\texttt{Qwen3.5-9B}} ($+5.95$ points on PersonaMemv3 and $+3.39$ on Mem-PAL over the strongest baselines) while remaining closely competitive with \textsc{GEPA} on \mbox{\texttt{Qwen3.8-27B}}. The subcategory breakdown in Appendix~\ref{app:subcategories} further reveals the source of these gains: on PersonaMemv3 with \mbox{\texttt{Qwen3.5-9B}}, \ourmethod{} simultaneously improves active preference application (51.40 versus 49.87 on Personalization) and context restraint (77.74 versus 66.88 on Restraint), while on MemoryCD with \mbox{\texttt{Qwen3.8-27B}} it substantially improves preference-driven item ranking (28.39 versus 24.39) alongside a minor trade-off on surface-level review summarization (12.34 versus 13.58). This indicates that learned harnesses do not merely retrieve more personal facts, but also regulate when personal context should be emphasized or withheld.

\paragraph{Targeted evidence selection decouples quality from deployment token cost.}
Table~\ref{tab:results} also shows that higher personalization quality does not require larger runtime contexts. Long-context prompting, persistent memory updates, and per-user prompt optimization incur substantial deployment overhead (typically 22k--64k tokens per query), yet frequently trail more selective context construction. By delegating history filtering to lightweight code and amortizing optimization across training users, \ourmethod{} achieves top scores with a compact deployment footprint, requiring only 2.07k--2.23k tokens per query on PersonaMemv3 with \mbox{\texttt{Qwen3.5-9B}} and \mbox{\texttt{GPT-5.6-Luna}} (less than half of \textsc{BM25}) and 4.46k--4.96k across all executors on MemoryCD. On Mem-PAL, \textsc{BM25} remains cheaper in tokens but sacrifices substantial accuracy, whereas \ourmethod{} reduces deployment tokens by more than $60\%$ relative to \textsc{Long Context}, \textsc{GEPA}, and \textsc{Rolling Memory}. Since Table~\ref{tab:results} combines setup and execution usage across different evaluation scopes, Section~\ref{sec:analysis_efficiency} further isolates runtime and scaling efficiency under strictly matched conditions.

\section{Analysis}
\label{sec:analysis}

\subsection{Model-Dependent Harnesses}
\label{sec:analysis_models}
\begin{figure}[!t]
\centering
\includegraphics[width=0.9\linewidth]{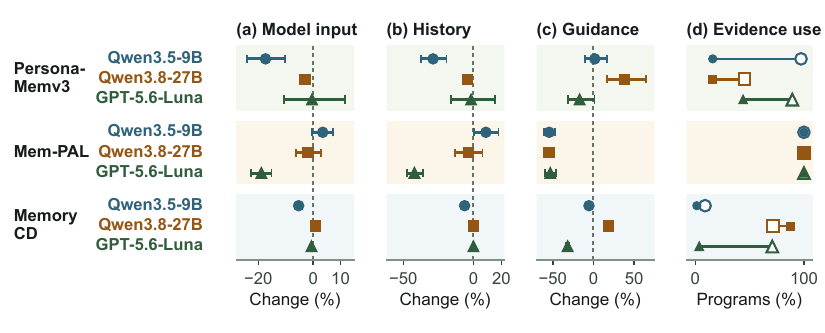}
\caption{Relative changes in model input, packed history, guidance length, and evidence-reading programs across nine runs, where open and filled symbols in (d) mark initial and trained programs. Optimization reduces model input in seven of nine configurations while tailoring context allocation to each task and executor.}
\label{fig:optimization_shifts}
\end{figure}

Figure~\ref{fig:optimization_shifts} tracks how cross-user optimization alters model input, packed history, answer guidance, and static evidence use across all nine runs. Across seven of nine configurations, optimization shrinks total model input while raising task accuracy; on PersonaMemv3 with \mbox{\texttt{Qwen3.5-9B}}, it delivers a $+7.96$-point gain while cutting packed history by nearly a third. Filtering distracting interactions therefore matters more than supplying longer prompts. Equally striking, the learned context allocation depends strongly on the answering model itself. Even when \mbox{\texttt{Qwen3.8-27B}} and \mbox{\texttt{GPT-5.6-Luna}} receive identical retrieved history on MemoryCD, optimization steers \mbox{\texttt{Qwen3.8-27B}} toward richer, evidence-conditioned guidance while shifting \mbox{\texttt{GPT-5.6-Luna}} toward concise, largely static preference cues. A similar split appears on PersonaMemv3 and Mem-PAL, where stronger models tolerate aggressive history compression while smaller models rely more heavily on explicit historical context. Effective harnesses thus co-adapt to the reasoning habits of each executor rather than following a single model-agnostic design (Appendix~\ref{app:model_dependent_analysis}, Table~\ref{tab:matched_start}; trajectories in Appendix~\ref{app:optimization_dynamics}).

\subsection{Generalization to Unseen User Clusters}
\label{sec:analysis_clusters}
\begin{figure}[!t]
\centering
\includegraphics[width=0.9\linewidth]{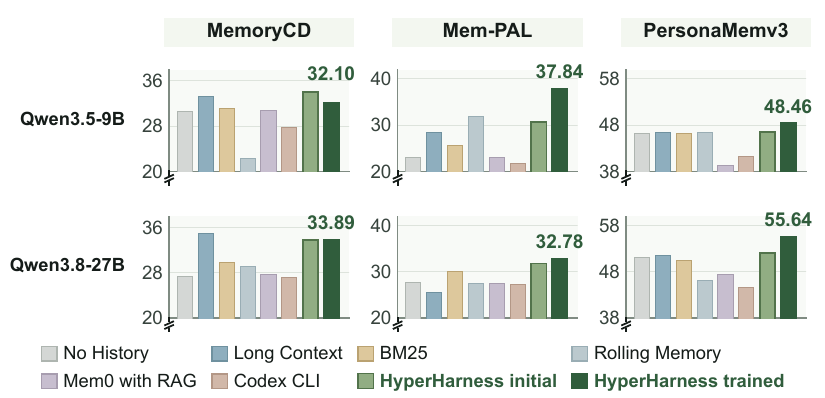}
\caption{Cluster-disjoint TEST scores with \mbox{\texttt{Qwen3.5-9B}} (top) and \mbox{\texttt{Qwen3.8-27B}} (bottom), with initial and trained \ourmethod{} highlighted in green. Vertical axes are truncated so bar-length ratios are not meaningful; full scores in Table~\ref{tab:cluster_generalization_details} show strong transfer on Mem-PAL and PersonaMemv3 alongside more limited transfer on MemoryCD.}
\label{fig:cluster_generalization}
\end{figure}

We further examine whether learned harness generation transfers to user groups whose preference profiles differ substantially from the training cohort. Using a fixed history-derived split where training and test users belong to disjoint clusters, we freeze the ten-round TRAIN-selected generator and evaluate it on held-out TEST clusters against six baselines (Figure~\ref{fig:cluster_generalization}; full scores in Table~\ref{tab:cluster_generalization_details} and cohort details in Appendix~\ref{app:cluster_details}). On Mem-PAL and PersonaMemv3, trained \ourmethod{} consistently improves over initialization and outperforms all baselines across both Qwen executors (by up to 5.99 points on Mem-PAL and 4.22 on PersonaMemv3). This transfer occurs because cross-user optimization learns domain-general \emph{procedural} habits, such as separating persistent preferences from incidental behavior and structuring targeted queries, rather than memorizing cluster-specific tastes. By contrast, transfer on MemoryCD is weaker, with both trained versions trailing \textsc{Long Context} and \mbox{\texttt{Qwen3.5-9B}} dipping below initialization. Because MemoryCD already imposes a cross-domain transfer to Home interactions, holding out entire user clusters compounds domain shift with population shift, making specialized retrieval rules more prone to overfitting. These results indicate that procedural transfer succeeds when evidence-selection principles remain invariant across user groups.

\subsection{Robustness to User History}
\label{sec:analysis_history}
\begin{figure}[!t]
\centering
\includegraphics[width=0.9\linewidth]{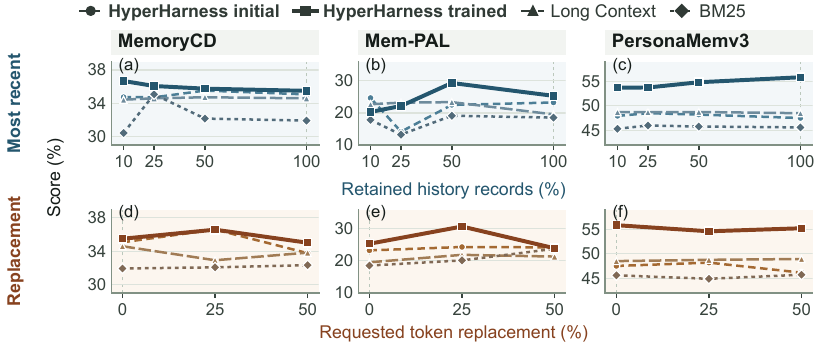}
\caption{History robustness under recent-record retention (top, blue) and irrelevant-text replacement (bottom, orange) with \mbox{\texttt{Qwen3.5-9B}}, 12 users per dataset, and one seed. Retention uses record fractions; replacement uses token fractions.}
\label{fig:history_robustness_main}
\end{figure}

Real-world interaction logs are frequently incomplete or noisy, so we evaluate \ourmethod{} with \mbox{\texttt{Qwen3.5-9B}} across 12 fixed users per dataset under two perturbations: truncating construction history to recent or random subsets, and replacing a fraction of original text tokens with irrelevant catalogue text alongside a matched deletion control that isolates distraction from missing evidence (Figure~\ref{fig:history_robustness_main}; controls in Appendix Figure~\ref{fig:history_robustness_controls}). Under severe sparsity, trained \ourmethod{} remains resilient: even with only the most recent 10\% of records, it stays well above full-history \textsc{Long Context} and \textsc{BM25} on PersonaMemv3 and slightly improves on MemoryCD, suggesting that concise recent interactions often suffice to infer useful retrieval rules while shedding older noise. Comparing replacement against matched deletion exposes a sharper asymmetry between missing evidence and active distraction. While trained harnesses largely absorb moderate noise on PersonaMemv3 and MemoryCD, injected distractors degrade Mem-PAL considerably more than deleting the same records (by 9.16 points at 50\% corruption), and intermediate retention levels fluctuate non-monotonically. For short preference-selection logs, filtering misleading records is therefore harder than handling missing ones.

\subsection{Efficiency Analysis}
\label{sec:analysis_efficiency}
\begin{figure}[!t]
\centering
\includegraphics[width=0.9\linewidth]{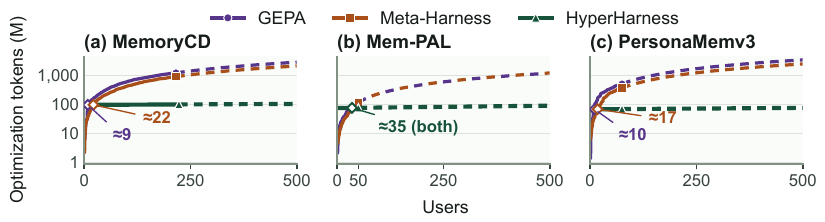}
\caption{Cumulative optimization token costs across users with \mbox{\texttt{Qwen3.5-9B}} and a 200-record history cap, where markers indicate the end of per-user sums, dashed lines extrapolate to 500 users, and diamonds mark crossover points. By amortizing shared training across users, \ourmethod{} scales at a much lower marginal cost than per-user search baselines.}
\label{fig:history200_token_demand}
\end{figure}

Practical personalization requires improving downstream quality without imposing prohibitive inference prompts or repeated per-user search costs. \ourmethod{} achieves this by structurally decoupling one-time cross-user training, single-pass per-user harness construction, and lightweight online execution (Table~\ref{tab:results}). In a matched MemoryCD comparison on \mbox{\texttt{Qwen3.5-9B}} across all 223 users and 2,676 requests, generated retrieval cuts mean answer-time tokens by 90.32\% relative to \textsc{Long Context} (3.15k versus 32.51k per query) while matching its task score. While this comparison isolates runtime execution from training cost, it shows that most raw history supplied to long-context models is redundant at query time and can be filtered programmatically before invoking the answering model. To examine multi-user scaling, Figure~\ref{fig:history200_token_demand} compares cumulative adaptation tokens under a 200-record history cap, scaling per-user optimization costs for \textsc{GEPA} and \textsc{Meta-Harness} to the 200-record budget alongside \ourmethod{}'s shared training and per-user construction costs. Per-user search methods (\textsc{GEPA} and \textsc{Meta-Harness}) accumulate steep marginal costs with every new user and frequently exceed the 32K input budget on \textsc{GEPA}. By contrast, \ourmethod{} pays a fixed training cost followed by lightweight per-user construction, crossing below both baselines after roughly 9--35 users for scalable deployment.

\subsection{Case Study}
\label{sec:analysis_case_study}
\begin{figure}[!t]
\centering
\includegraphics[width=0.7\linewidth]{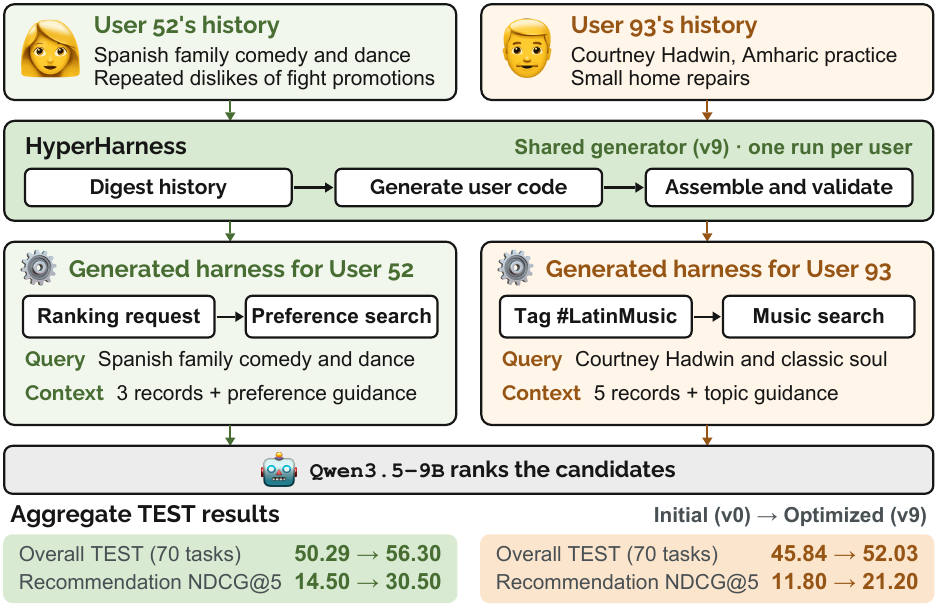}
\caption{Harness construction, recommendation traces, and aggregate results for two unseen PersonaMemv3 users.}
\label{fig:case_study}
\end{figure}

To understand how a single frozen generator adapts execution logic to heterogeneous users, Figure~\ref{fig:case_study} traces initial and optimized \ourmethod{} programs built for two unseen PersonaMemv3 users with the lowest and highest shares of short-term preference labels, both executed by \mbox{\texttt{Qwen3.5-9B}}. From the same construction pipeline, the optimized generator synthesizes distinct control flows matched to each user's history instead of applying a uniform template: for User~52, whose log reflects concentrated tastes (Spanish family comedy and dance alongside dislikes of fight promotions), the optimized generator replaces the initial harness's skipped retrieval with a ranking-triggered preference search that supplies three records and preference guidance; for User~93, whose log spans diverse interests (music, Amharic practice, and home repairs), the same generator instead matches candidate cues such as \texttt{\#LatinMusic} to trigger a targeted music search (Courtney Hadwin and classic soul) with five records and topic guidance. Across all 70 TEST requests per user, these tailored procedures lift overall scores by about six points (50.29 to 56.30 and 45.84 to 52.03) and nearly double mean recommendation NDCG@5 (14.50 to 30.50 and 11.80 to 21.20). On the two illustrated slates, however, NDCG@5 remains zero under both versions because the first search exhausts the history budget and hard negatives remain in the top five. These traces show that cross-user optimization improves personalization mainly by turning user history into tailored retrieval triggers, queries, and guidance logic rather than expanding raw context, while single-shot retrieval budgets remain a limiting factor when candidate slates contain closely competing distractors.

\section{Related Work}
\label{sec:related}
\paragraph{LLM personalization.}
Existing LLM personalization methods adapt context, model parameters, alignment objectives, or agent behavior \citep{zhang2025personalizationsurvey,liu2025personalizedsurvey,guan2025alignment}. Prompt- and context-based approaches incorporate user data directly into the input context, including raw user history as few-shot examples \citep{dai2023uncovering,wang2023learning,kang2023llms,kim-yang-2025-shot}, history retrieval to handle long logs \citep{salemi2024lamp,mysore2024pearl,salemi2024retrieval}, and summarized textual user profiles \citep{richardson2023summarization,sun2024persona,dong-etal-2024-llm,tan2025perrecbench,zhang2024gpg}; subsequent work enhances these pipelines with planning, reasoning, and utility-driven profile optimization \citep{salemi2025lampqa,salemi2025restpg,gao2025langptune,tan2026blue}. Parameter-based approaches embed user preferences in lightweight model components, spanning per-user PEFT adapters \citep{tan2024oppu}, collaborative parameter sharing \citep{tan2024perpcs}, group-level adaptation \citep{zhang2025proper}, generated user adapters \citep{tan2026profiletopeft}, and learned user embeddings \citep{ning2025user}. A complementary line targets personalized alignment via post-hoc parameter merging \citep{jang2023personalized}, personalized RLHF \citep{li2024personalizedrlhf,park2024principled}, customized or factorized reward models \citep{cheng2023deserves,bose2025lore,shenfeld2025language}, black-box model factorization \citep{zhuang2024hydra}, and interactive conversational settings \citep{zhao2025personalens,wu-etal-2025-aligning}. More recently, agent personalization has extended adaptation to memory, planning, tool use, preference-aware skills, and harness evolution \citep{zhang2026personaagent,mazzolenis2025warpp,bose2026cape,liang2026pahf,gan2026localharness,jin2026fable,gan2026harnessevolution}. Rather than updating executor weights or relying on a fixed retrieval-and-prompting template, \ourmethod{} learns a shared generator of executable user-specific procedures so history shapes both retrieved evidence and the code that applies it.

\paragraph{Automated agent and harness design.}
Automatic optimization has progressed from prompts and agent workflows \citep{khattab2024dspy,agrawal2026gepa,hu2025adas,zhang2025aflow} to evolving memories and skills \citep{ouyang2026skillos,ouyang2026reasoningbank,wei2026evoharness,xu2026hyperskill}. Another line improves agent code and meta-level modification procedures recursively, including STOP, SICA, the Darwin G\"odel Machine, Hyperagents, MetaSkill-Evolve, and MCE \citep{zelikman2024stop,robeyns2025sica,zhang2025dgm,zhang2026hyperagents,wang2026metaskill,ye2026metacontext}. Recent work extends automated search directly to end-to-end executable harnesses: offline methods evolve harnesses with execution feedback, observability signals, or research knowledge \citep{lee2026metaharness,yang2026scholarevolve,lin2026agentic,chen2026harnessx,zhang2026darwinx,zhang2026harnesscompass}, while related systems train harness editors and generators from rewards or diagnostic search experience \citep{shao2026harnessr1,zhang2026jitagent,huang2026memoharness}. Online and test-time approaches further refine harnesses across deployment streams and continual episodes \citep{lee2026rhi,wang2026rethinking,nie2026tthe,liu2026adaptiveah,karten2026continual,kang2026harness,zhou2026hsi,xu2026harness2}. \ourmethod{} applies harness optimization to a shared construction procedure across users: task-scored search happens during cross-user training, so the frozen generator synthesizes each new user's harness directly from history without per-user scored search at deployment.

\section{Conclusion}
We introduced \ourmethod{}, a framework for learning to generate personalized agent harnesses from user histories. By optimizing a shared construction procedure with downstream feedback, it produces user-specific retrieval and guidance code without per-user task-scored search at deployment. Experiments show competitive performance across three benchmarks and reveal that useful adaptation can come from changing how context is selected and interpreted. The findings support executable harness generation as a practical direction for personalization, with transfer and history sensitivity remaining important considerations.

\section*{Limitations}
Our experiments cover three benchmarks under bounded retrieval-and-guidance interfaces and do not test unrestricted agent workflows. Several cross-method comparisons span slightly different evaluation subsets or task masks, and the robustness study evaluates 12 users per dataset with a single perturbation seed. Both harness optimization and construction are stochastic, and our case studies examine end-to-end traces without ablating individual code rules. Finally, cost comparisons reflect different workload scopes, and deploying generated programs in practice requires strict access controls, validation, and privacy safeguards over user histories.

\bibliography{references}
\bibliographystyle{unsrtnat}

\clearpage
\appendix
\section{Implementation Details}
\label{app:implementation}
\paragraph{Models and optimization.}
We use \texttt{gpt-6-sol} for harness construction and generator optimization, and \texttt{Qwen3.5-9B}, \texttt{Qwen3.8-27B}, or \texttt{GPT-5.6-Luna} for answer execution. Each run evaluates an initial generator and up to ten optimization rounds, then freezes a TRAIN-selected version for TEST. Construction and optimization use medium reasoning effort by default, with low construction effort for the \mbox{\texttt{Qwen3.5-9B}} MemoryCD configuration and task-aware initialization on Mem-PAL. Independent \texttt{Qwen} replications start from fresh initial programs and reference scaffolds and are reported separately.

\paragraph{Candidate selection.}
A candidate is normally adopted when it improves the equal-user mean on the TRAIN panel. The optimizer receives the current specification, generated user programs, task scores, and diagnostic cases showing gains and regressions relative to the incumbent. Candidate evaluation generates new personal programs, so the measured change includes construction variability as well as changes to the shared procedure. Construction checks test interface compliance and executable validity with permitted history and synthetic requests; their repair feedback is separate from task-quality feedback. For \mbox{\texttt{GPT-5.6-Luna}} on MemoryCD, candidate acceptance additionally requires gains in at least three of five fixed TRAIN groups and a per-task regression check against the initial generator.

\paragraph{Execution and baselines.}
The input cap is 32,768 tokens. Output caps are 512 for Mem-PAL and 4,096 for MemoryCD; PersonaMemv3 uses its official task-specific limits. Mem-PAL and MemoryCD harnesses use a single final-answer call per request. \texttt{Qwen} disables implicit thinking and uses local structured decoding, while \mbox{\texttt{GPT-5.6-Luna}} uses hosted decoding with reasoning effort \texttt{none}. \textsc{Long Context} packs recent complete events; \textsc{BM25} retrieves ten records per query; \textsc{Rolling Memory} updates every 20 events with a 512-token memory cap; and \textsc{Mem0 with RAG} uses the corresponding executor for updates and \texttt{MiniLM} top-10 retrieval without reranking. \textsc{Codex CLI} permits at most 12 model calls and 600 seconds per answer. We also compare \textsc{GEPA} \citep{agrawal2026gepa} and \textsc{Meta-Harness} \citep{lee2026metaharness}; \textsc{No History} is the non-personalized reference. Methods retain their reported budgets, so equal input caps do not imply equal end-to-end compute. Appendix~\ref{app:prompts_and_harness} gives the prompt templates and the executed code examples.

\paragraph{Evaluation metrics and deployment cost.}
We compute each user's benchmark score and average equally across users. PersonaMemv3 uses the official rule-based and \texttt{gpt-5.5} evaluators with lifecycle-pair aggregation (0--100 scale), following each user's designated later-half query split. Mem-PAL uses $50\times(\text{positive hits}-\text{negative hits})$ on its English machine-translated solution-selection subset ($-100$ to $100$ scale). MemoryCD equally weights NDCG@5 for item ranking, normalized rating accuracy $1-\mathrm{MAE}/4$, and ROUGE-L for review titles and review text (0--100 scale). Chronological evaluations admit later observations only when the task protocol makes them available, whereas MemoryCD uses fixed source-domain history. Deployment tokens $T_{\mathrm{deploy}}$ are the equal-user mean of $(T_{\mathrm{setup},u}+T_{\mathrm{exec},u})/Q_u$ in thousands per scheduled query, where setup includes memory or harness construction and execution includes answer calls and subsequent memory updates; judging and shared generator training are excluded. Because methods can cover slightly different evaluation subsets, \mbox{\texttt{GPT-5.6-Luna}} uses filtered task masks on MemoryCD and Mem-PAL, and \ourmethod{} deployment tokens on Mem-PAL with \mbox{\texttt{Qwen3.5-9B}} are measured from a separate replication.

\section{Data Statistics}
\label{app:data_statistics}
\begingroup
\setlength{\intextsep}{3pt}
\begin{wrapfigure}{r}{0.63\linewidth}
\centering
\captionof{table}{User and scheduled item counts for the main experimental splits.}
\label{tab:data_statistics}
\begingroup
\small
\setlength{\tabcolsep}{2pt}
\hhtablelayout
\begin{tabularx}{\linewidth}{l*{6}{H}}
\toprule
\rowcolor{hhTableGroup}& \multicolumn{2}{c}{\textbf{PersonaMemv3}} & \multicolumn{2}{c}{\textbf{Mem-PAL}} & \multicolumn{2}{c}{\textbf{MemoryCD}} \\
\cmidrule(lr){2-3}\cmidrule(lr){4-5}\cmidrule(lr){6-7}
\rowcolor{hhTableHead}\textbf{Measure} & Train & Test & Train & Test & Train & Test \\
\midrule
\rowcolor{hhTableStripe}Users & 25 & 75 & 50 & 50 & 100 & 223 \\
\rowcolor{white}Items & 1,860 & 5,700 & 418 & 408 & 1,200 & 2,676 \\
\rowcolor{hhTableStripe}Mean items per user & 74.40 & 76.00 & 8.36 & 8.16 & 12.00 & 12.00 \\
\bottomrule
\end{tabularx}
\endgroup
\end{wrapfigure}

Table~\ref{tab:data_statistics} summarizes the user-disjoint training and test splits. MemoryCD has the largest test cohort (223 users). PersonaMemv3 has the most test items (5,700), averaging 76 per user compared with 12 on MemoryCD and 8.16 on Mem-PAL. We count each scheduled item once across methods and models.
\par
\ifnum\value{WF@wrappedlines}>1\relax
  \vspace{\dimexpr\value{WF@wrappedlines}\baselineskip-\baselineskip\relax}
\fi
\WFclear
\endgroup

\section{Supplementary Analyses}
\label{app:analysis_protocols}
\subsection{Model-Dependent Harnesses}
\label{app:model_dependent_analysis}
\begin{table}[!t]
\centering
\begin{minipage}{0.90\linewidth}
\centering
\caption{Initial and optimized HyperHarness results. Lengths and input tokens are per query. MemoryCD TEST scores and \texttt{Qwen3.5-9B} measures use the matched-initialization subset; remaining MemoryCD context measures come from the full main runs.}
\label{tab:matched_start}
\begingroup
\normalsize
\setlength{\tabcolsep}{2pt}
\hhtablelayout
\begin{tabularx}{\linewidth}{l*{6}{H}}
\toprule
\rowcolor{hhTableGroup}\textbf{MemoryCD} & \multicolumn{2}{c}{\texttt{Qwen3.5-9B}} & \multicolumn{2}{c}{\texttt{Qwen3.8-27B}} & \multicolumn{2}{c}{\texttt{GPT-5.6-Luna}} \\
\cmidrule(lr){2-3}\cmidrule(lr){4-5}\cmidrule(lr){6-7}
\rowcolor{hhTableHead}\textbf{Measure} & Initial & Optimized & Initial & Optimized & Initial & Optimized \\
\midrule
\rowcolor{hhTableStripe}TEST score & 35.52 & 36.97 & 35.26 & 37.88 & 38.72 & 39.43 \\
\rowcolor{white}History characters (k) & 10.09 & 9.93 & 10.12 & 10.12 & 10.12 & 10.12 \\
\rowcolor{hhTableStripe}Guidance characters & 486.3 & 251.5 & 485.6 & 577.0 & 478.8 & 326.3 \\
\rowcolor{white}Reads evidence (\% programs) & 91.7 & 0.0 & 71.7 & 87.4 & 70.9 & 3.6 \\
\rowcolor{hhTableStripe}Input tokens & 3,642.12 & 3,569.62 & 3,623 & 3,654 & 3,472 & 3,449 \\
\bottomrule
\end{tabularx}
\par\smallskip
\begin{tabularx}{\linewidth}{l*{6}{H}}
\toprule
\rowcolor{hhTableGroup}\textbf{Mem-PAL} & \multicolumn{2}{c}{\texttt{Qwen3.5-9B}} & \multicolumn{2}{c}{\texttt{Qwen3.8-27B}} & \multicolumn{2}{c}{\texttt{GPT-5.6-Luna}} \\
\cmidrule(lr){2-3}\cmidrule(lr){4-5}\cmidrule(lr){6-7}
\rowcolor{hhTableHead}\textbf{Measure} & Initial & Optimized & Initial & Optimized & Initial & Optimized \\
\midrule
\rowcolor{hhTableStripe}TEST score & 28.72 & 32.58 & 30.08 & 30.16 & 29.97 & 30.36 \\
\rowcolor{white}History characters (k) & 11.59 & 12.62 & 11.52 & 11.12 & 10.66 & 6.17 \\
\rowcolor{hhTableStripe}Guidance characters & 631.9 & 287.8 & 630.6 & 284.2 & 642.3 & 300.3 \\
\rowcolor{white}Reads evidence (\% programs) & 100.0 & 100.0 & 100.0 & 100.0 & 100.0 & 100.0 \\
\rowcolor{hhTableStripe}Input tokens & 7,487.70 & 7,746.40 & 7,466.75 & 7,336.04 & 6,347.84 & 5,146.47 \\
\bottomrule
\end{tabularx}
\par\smallskip
\begin{tabularx}{\linewidth}{l*{6}{H}}
\toprule
\rowcolor{hhTableGroup}\textbf{PersonaMemv3} & \multicolumn{2}{c}{\texttt{Qwen3.5-9B}} & \multicolumn{2}{c}{\texttt{Qwen3.8-27B}} & \multicolumn{2}{c}{\texttt{GPT-5.6-Luna}} \\
\cmidrule(lr){2-3}\cmidrule(lr){4-5}\cmidrule(lr){6-7}
\rowcolor{hhTableHead}\textbf{Measure} & Initial & Optimized & Initial & Optimized & Initial & Optimized \\
\midrule
\rowcolor{hhTableStripe}TEST score & 46.55 & 54.51 & 51.50 & 52.83 & 56.06 & 58.59 \\
\rowcolor{white}History characters (k) & 5.55 & 3.95 & 13.05 & 12.48 & 5.18 & 5.09 \\
\rowcolor{hhTableStripe}Guidance characters & 289.2 & 294.2 & 307.2 & 425.5 & 285.8 & 237.1 \\
\rowcolor{white}Reads evidence (\% programs) & 97.3 & 16.0 & 45.3 & 16.0 & 89.3 & 44.0 \\
\rowcolor{hhTableStripe}Input tokens & 2,062.38 & 1,702.76 & 3,942.84 & 3,824.27 & 1,882.60 & 1,875.46 \\
\bottomrule
\end{tabularx}
\endgroup
\end{minipage}
\end{table}

\paragraph{Experimental setup and evaluation.}
The Mem-PAL runs share the initial generator, construction and optimizer settings, TRAIN users, ten-round budget, and evaluation panel at both stages. Selected rounds are 1, 2, and 10 for \mbox{\texttt{Qwen3.5-9B}}, \mbox{\texttt{Qwen3.8-27B}}, and \mbox{\texttt{GPT-5.6-Luna}}. In the MemoryCD matched-initialization comparison, all executors start from the exact same initial generator and are evaluated on identical TEST queries at both stages. \mbox{\texttt{Qwen3.5-9B}} uses 12 TRAIN users and three optimization rounds and selects round 2. \mbox{\texttt{Qwen3.8-27B}} and \mbox{\texttt{GPT-5.6-Luna}} reuse their full training histories and select rounds 10 and 5, respectively. Figure~\ref{fig:optimization_shifts} reports the main MemoryCD runs, including a six-round run for \mbox{\texttt{Qwen3.5-9B}}. On PersonaMemv3, the initial and optimized results are paired within each executor. For context lengths, we sum characters over answer calls per query, average queries within each user, and weight users equally. Evidence use measures the share of programs reading the evidence argument.

\paragraph{Changes in context use.}
In the matched-initialization MemoryCD comparison, training reduces guidance length by 48.3\% for \mbox{\texttt{Qwen3.5-9B}} and by 26.7\% for \mbox{\texttt{GPT-5.6-Luna}}, with the share of programs reading the evidence argument falling from 91.7\% and 66.7\%, respectively, to zero, whereas \mbox{\texttt{Qwen3.8-27B}} expands guidance by 16.3\% and raises the share of programs reading the evidence argument to 83.3\%. On Mem-PAL, guidance shrinks by 53--55\% across all three models, yet packed history shifts in opposite directions, increasing by 8.9\% for \mbox{\texttt{Qwen3.5-9B}} while decreasing by 3.5\% for \mbox{\texttt{Qwen3.8-27B}} and 42.1\% for \mbox{\texttt{GPT-5.6-Luna}}. On PersonaMemv3, relative to each executor's own initialization, \mbox{\texttt{Qwen3.8-27B}} uses 38.5\% more guidance alongside a 4.4\% shorter history, while \mbox{\texttt{Qwen3.5-9B}} keeps guidance length nearly unchanged ($+1.7\%$) and cuts packed history by 28.8\%.

We measure PersonaMemv3 history and guidance lengths for all three models by replaying the frozen programs against the same cached request sequence. Each model has one optimization trajectory per dataset. Differences can reflect both construction variability and the selected round.

\paragraph{Selected MemoryCD harnesses.}
\mbox{\texttt{Qwen3.8-27B}} and \mbox{\texttt{GPT-5.6-Luna}} receive identical history text on all 2,436 matched queries. \mbox{\texttt{GPT-5.6-Luna}} uses 43.3\% fewer guidance characters per query.

\subsection{Generalization to Unseen User Clusters}
\label{app:cluster_details}
\begin{table}[!t]
\centering
\begin{minipage}{0.93\linewidth}
\centering
\caption{TEST scores on the fixed cluster-disjoint split. Bold and underline identify the best and second-best history-aware methods. Gray marks the \textsc{No History} reference.}
\label{tab:cluster_generalization_details}
\begingroup
\footnotesize
\setlength{\tabcolsep}{1.5pt}
\renewcommand{\arraystretch}{1.12}
\hhtablelayout
\begin{tabularx}{\linewidth}{>{\raggedright\arraybackslash}X*{6}{r}}
\toprule
\rowcolor{hhTableSubhead}& \multicolumn{2}{c}{MemoryCD} & \multicolumn{2}{c}{Mem-PAL} & \multicolumn{2}{c}{PersonaMemv3} \\
\cmidrule(lr){2-3}\cmidrule(lr){4-5}\cmidrule(lr){6-7}
\rowcolor{hhTableHead}\textbf{Method} & \mbox{\fontsize{8}{9}\selectfont\texttt{Qwen3.5-9B}} & \mbox{\fontsize{8}{9}\selectfont\texttt{Qwen3.8-27B}} & \mbox{\fontsize{8}{9}\selectfont\texttt{Qwen3.5-9B}} & \mbox{\fontsize{8}{9}\selectfont\texttt{Qwen3.8-27B}} & \mbox{\fontsize{8}{9}\selectfont\texttt{Qwen3.5-9B}} & \mbox{\fontsize{8}{9}\selectfont\texttt{Qwen3.8-27B}} \\
\midrule
\rowcolor{hhTableStripe}\textcolor{hhTableMuted}{\textsc{No History}} & \textcolor{hhTableMuted}{30.62} & \textcolor{hhTableMuted}{27.41} & \textcolor{hhTableMuted}{23.10} & \textcolor{hhTableMuted}{27.67} & \textcolor{hhTableMuted}{46.09} & \textcolor{hhTableMuted}{51.01} \\
\rowcolor{white}\textsc{Long Context} & \underline{33.23} & \textbf{34.94} & 28.38 & 25.44 & \underline{46.51} & 51.42 \\
\rowcolor{hhTableStripe}\textsc{BM25} & 31.15 & 29.87 & 25.61 & 30.04 & 46.14 & 50.39 \\
\rowcolor{white}\textsc{Rolling Memory} & 22.33 & 29.16 & \underline{31.86} & 27.37 & 46.43 & 46.11 \\
\rowcolor{hhTableStripe}\textsc{Mem0 with RAG} & 30.80 & 27.64 & 22.92 & 27.41 & 39.27 & 47.33 \\
\rowcolor{white}\textsc{Codex CLI} & 27.85 & 27.16 & 21.61 & 27.25 & 41.18 & 44.66 \\
\rowcolor{hhTableAccentLight}HyperHarness (initial) & \textbf{34.08} & 33.82 & 30.68 & \underline{31.81} & 46.44 & \underline{51.98} \\
\rowcolor{hhTableAccent}HyperHarness (optimized) & 32.10 & \underline{33.89} & \textbf{37.84} & \textbf{32.78} & \textbf{48.46} & \textbf{55.64} \\
\bottomrule
\end{tabularx}
\par\smallskip
{\scriptsize HyperHarness rows report the original ten-round experiment. Methods use different compute budgets.\par}
\endgroup
\end{minipage}
\end{table}

TRAIN cohorts contain 56 MemoryCD, 24 Mem-PAL, and 12 PersonaMemv3 users. Table~\ref{tab:cluster_generalization_details} reports TEST scores for Figure~\ref{fig:cluster_generalization} from the ten-round HyperHarness experiment; methods retain their compute budgets.

\begin{table}[htbp]
\centering
\begin{minipage}{0.70\linewidth}
\centering
\caption{Deployment-model rollout tokens during optimization (input plus output).}
\label{tab:optimization_rollout_usage}
\begingroup
\normalsize
\setlength{\tabcolsep}{2pt}
\hhtablelayout
\begin{tabularx}{\linewidth}{llrrH}
\toprule
\rowcolor{hhTableGroup}& & \multicolumn{3}{c}{\textbf{Rollout tokens (M)}} \\
\cmidrule(lr){3-5}
\rowcolor{hhTableHead}\textbf{Benchmark} & \textbf{Model} & \textbf{HyperHarness} & \textbf{Meta-Harness} & \textbf{GEPA} \\
\midrule
\rowcolor{hhTableStripe} & \texttt{Qwen3.5-9B} & 45.67 & 322.95 & 502.92 \\
\rowcolor{hhTableStripe}\textbf{PersonaMemv3} & \texttt{Qwen3.8-27B} & 81.07 & 1,014.91 & 427.10 \\
\rowcolor{hhTableStripe} & \texttt{GPT-5.6-Luna} & 39.82 & 444.94 & 321.59 \\
\midrule
\rowcolor{hhTableStripe} & \texttt{Qwen3.5-9B} & 39.96 & 100.03 & 118.09 \\
\rowcolor{hhTableStripe}\textbf{Mem-PAL} & \texttt{Qwen3.8-27B} & 37.24 & 330.01 & 99.41 \\
\rowcolor{hhTableStripe} & \texttt{GPT-5.6-Luna} & 23.98 & 200.83 & 84.09 \\
\midrule
\rowcolor{hhTableStripe} & \texttt{Qwen3.5-9B} & 56.66 & 818.00 & 1,205.71 \\
\rowcolor{hhTableStripe}\textbf{MemoryCD} & \texttt{Qwen3.8-27B} & 56.85 & 2,121.89 & 1,076.17 \\
\rowcolor{hhTableStripe} & \texttt{GPT-5.6-Luna} & 43.73 & 838.85 & 1,203.01 \\
\bottomrule
\end{tabularx}
\endgroup
\end{minipage}
\end{table}

\subsection{Optimization Dynamics}
\label{app:optimization_dynamics}
\begin{figure}[!t]
\centering
\includegraphics[width=0.9\linewidth]{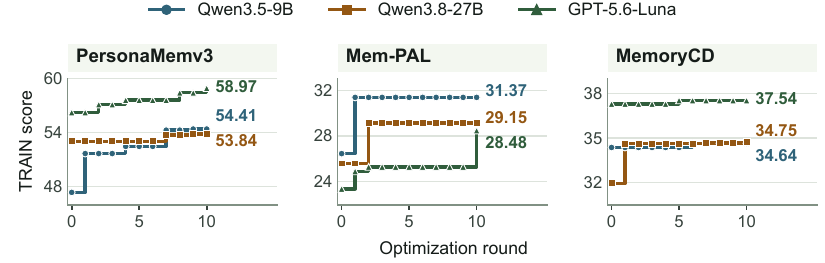}
\caption{TRAIN scores of the incumbent generator across optimization rounds. Plateaus indicate no adopted improvement. Panels use different vertical scales; these curves measure training progress, not held-out gains.}
\label{fig:training_progress}
\end{figure}

Figure~\ref{fig:training_progress} tracks the adopted generator's equal-user TRAIN score from initialization through ten search rounds. Flat segments indicate an unchanged incumbent. With \mbox{\texttt{GPT-5.6-Luna}}, PersonaMemv3 rises from 56.26 to 58.97, Mem-PAL plateaus after round 2 to 28.48 at round 10, and MemoryCD moves from 37.30 to 37.54.

With \mbox{\texttt{GPT-5.6-Luna}}, optimization gains 2.71 TRAIN points and 1.26 paired TEST points on PersonaMemv3 (25 TRAIN users), and 5.10 TRAIN and 1.29 TEST points on Mem-PAL. On MemoryCD, adding group-level stability gating yields 0.28 TRAIN and 0.46 TEST gains, compared with 1.05 TRAIN and 0.07 TEST under unconstrained mean-gain selection, though evaluation panels also differ across runs.

\subsection{Robustness to User History}
\label{app:history_robustness}
\begin{figure}[!t]
\centering
\includegraphics[width=0.9\linewidth]{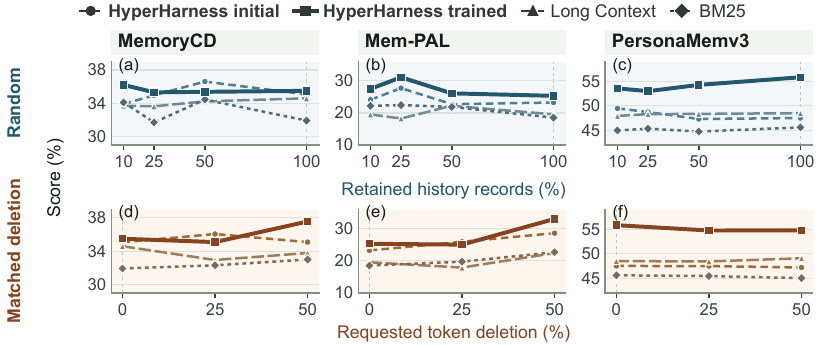}
\caption{Random-retention and matched-deletion controls for Figure~\ref{fig:history_robustness_main}, with the same users, generators and dataset-specific scales. Retention matches record counts; deletion uses the same whole records as replacement, so token fractions are approximate.}
\label{fig:history_robustness_controls}
\end{figure}

\paragraph{Setting.}
We use 12 hash-selected TEST users per benchmark and their questions: 144 on MemoryCD, 109 on Mem-PAL and 885 on PersonaMemv3. \mbox{\texttt{Qwen3.5-9B}} executes the initial generator and the frozen TRAIN-selected generators on each benchmark, with no TEST-based update. Construction uses \texttt{gpt-6-sol}, at low effort for MemoryCD/Mem-PAL and medium effort for PersonaMemv3; native scoring includes the official \texttt{gpt-5.5} judge where required. The Mem-PAL generator evaluated here is distinct from the main-table run.

\paragraph{Perturbations.}
Recent/random retention keeps $\lceil fN\rceil$ of $N$ construction-history records for $f\in\{0.1,\allowbreak 0.25,\allowbreak 0.5,\allowbreak 1\}$; retained records remain chronological. Replacement/deletion pairs select the same whole records to approximate 25\% or 50\% of original event-text tokens. Replacement substitutes non-personal museum, geological and observatory catalogue text and preserves the total event-text token count; metadata and model-input lengths may differ. Construction and retrieval use the same perturbed history. Only the construction prefix changes: later observations remain available on Mem-PAL and PersonaMemv3, while MemoryCD uses fixed source history.

\paragraph{Aggregation.}
We use a single fixed random seed across all 11 perturbation conditions and four methods. Full-history controls are shared across settings. We use each benchmark's scoring procedure to aggregate scores within users, then average equally across users. Each condition constructs a fresh harness, so differences can reflect both history sensitivity and construction variability.

\subsection{Connection to Hypernetworks}
\label{app:hypernetwork_discussion}
\begin{table}[!t]
\centering
\caption{Conceptual comparison between neural hypernetworks for LLM personalization and \ourmethod{}. Both paradigms amortize adaptation across users via a shared conditional generator, but they operate in fundamentally different output, optimization, and execution spaces.}
\label{tab:hypernetwork_comparison}
\begingroup
\small
\setlength{\tabcolsep}{5pt}
\hhtablelayout
\begin{tabularx}{\linewidth}{l >{\raggedright\arraybackslash}X >{\raggedright\arraybackslash}X}
\toprule
\rowcolor{hhTableHead}\textbf{Dimension} & \textbf{Neural Hypernetworks (e.g., PEFT Generation)} & \textbf{\ourmethod{} (Harness Generation)} \\
\midrule
\rowcolor{hhTableStripe}Conditioning input & Task embedding, prompt description, or textual user profile/history & Permitted raw user interaction history $\mathcal{H}_u^{\mathrm{con}}$ \\
\rowcolor{white}Generated artifact & Continuous weights or low-rank adapters ($\Delta\theta_u$) & Discrete executable retrieval and guidance code ($h_u$) \\
\rowcolor{hhTableStripe}Generator parameterization & Continuous neural weights trained end-to-end by gradient descent & Discrete prompts, code, and resources $S$ driving a frozen builder LLM \\
\rowcolor{white}Training signal & Backpropagated downstream loss through generated weights & Derivative-free cross-user optimization from downstream task scores \\
\rowcolor{hhTableStripe}Executor requirements & White-box access to a specific open-weight checkpoint and architecture & Frozen open-weight or hosted black-box executor via API/context \\
\rowcolor{white}Mechanism of adaptation & Modulates internal hidden representations and output distributions & Controls memory indexing, multi-query retrieval, filtering, and prompt guidance \\
\rowcolor{hhTableStripe}Auditability \& safety & Opaque continuous weight deltas; difficult to inspect per rule & Human-readable code subject to host validation and access controls \\
\bottomrule
\end{tabularx}
\endgroup
\end{table}

\ourmethod{} is conceptually inspired by neural hypernetworks \citep{ha2017hypernetworks} and hypernetwork-based parameter adaptation for LLMs \citep{phang2023hypertuning,ivison2023hint,tan2026profiletopeft}. Below we summarize the shared design principles and key differences between the two paradigms (Table~\ref{tab:hypernetwork_comparison}).

\paragraph{Shared principles.}
Both neural hypernetworks and \ourmethod{} address the tension between per-user specialization and deployment cost through \emph{amortized conditional generation}. Rather than running iterative fine-tuning or search separately for each unseen user at test time, both learn a single shared meta-generator across a training population and condition it on user-specific inputs to synthesize a personalized artifact in a single pass. Both also avoid ground-truth target artifacts: hypernetworks do not need supervised reference adapters when trained end-to-end on downstream task loss, and \ourmethod{} similarly avoids ground-truth reference harnesses by optimizing the shared generator $G_S$ directly against downstream answering quality across training users.

\paragraph{Key differences and complementarity.}
Despite this shared formulation, the two paradigms differ along four main axes. First, in \emph{output space and parameterization}, neural hypernetworks are differentiable models that map conditioning vectors into continuous parameter deltas ($\Delta\theta_u$), whereas \ourmethod{} parameterizes its generator with discrete prompts, code, and supporting specifications $S$ and generates executable programs $h_u$ via a frozen construction LLM. Second, in \emph{optimization}, neural hypernetworks rely on gradient backpropagation through generated weights, whereas \ourmethod{} uses derivative-free cross-user optimization driven by task scores and diagnostic execution traces. Third, in \emph{executor compatibility}, parameter-generating hypernetworks require white-box access to open-weight checkpoints and are coupled to a fixed model architecture, whereas \ourmethod{} keeps the answering model completely frozen and works uniformly across open-weight and hosted black-box models (such as \mbox{\texttt{GPT-5.6-Luna}}). Fourth, in \emph{adaptation mechanism and inspectability}, weight generation steers internal model representations with minimal extra input tokens at inference time, while harness generation adapts external control flow, query formulation, and evidence filtering in human-readable code that can be inspected, sandboxed, and validated by the host (Section~\ref{sec:analysis_case_study}). Consequently, the two approaches are complementary: \ourmethod{} is especially suited to long-horizon episodic memory selection and black-box deployment, whereas parameter-space hypernetworks could be combined with generated harnesses to jointly adapt both external context selection and internal model weights when open checkpoints are available.

\section{Detailed Subcategory Results}
\label{app:subcategories}
Subcategory results use the same run, users and evaluation mask as their corresponding Overall entries. Scores are aggregated within user and then equally across users with an eligible scoring unit; Overall retains each benchmark's original aggregation, rather than averaging the new category columns. Deployment tokens are per scheduled query, not sums of category means.

\paragraph{PersonaMemv3.}
We follow the published task taxonomy \citep{jiang2026personamemv3}, combining agentic and proactive tasks. \emph{Personalization} covers personal responses, location-aware recommendations, explicit/implicit personal QA and updated preferences; \emph{Recommendation} covers rankings, earlier directives, preference lifecycles and chatbot suggestions; \emph{Restraint} contains the six \texttt{over\_personalization\_*} tasks; and \emph{Agentic \& Proactive} covers cross-app actions, refinding, digests, alerts, reactions and intervention control. Silence/intervention tasks remain in the proactive group, and the official lifecycle-pair aggregation is preserved.

\paragraph{Mem-PAL.}
The 408 solution-selection requirements are assigned to \emph{Work \& Study}, \emph{Health \& Wellness}, \emph{Family \& Social}, or \emph{Daily Life \& Leisure}, informed by PAL-Set's domains \citep{huang2026mempal}. This is a reporting rubric, not an official task split. Each public requirement was mapped to one domain based on its main goal using an automated LLM classifier without access to model predictions or query scores; the mapping was frozen before grouped scoring and shared across all methods, and all domains retain the signed selection metric.

\paragraph{MemoryCD.}
The four official tasks \citep{zhang2026memorycd} are Item Ranking (NDCG@5), Rating Prediction ($1-\mathrm{MAE}/4$), Review Summarization (title ROUGE-L), and Review Generation (text ROUGE-L), each scaled by 100. They share the benchmark's source-history-to-Home transfer split and original four-task macro-average.

\begin{table}[!t]\centering
\caption{TEST subcategory scores, Overall scores and deployment tokens (thousands per scheduled query). Bold and underline mark the best and second-best values for each metric among personalized methods. Gray indicates the \textsc{No History} reference. NR denotes unreported values. Overall and deployment values reproduce Table~\ref{tab:results}.}
\label{tab:subcategory_results}
\begingroup
\footnotesize
\setlength{\tabcolsep}{2pt}
\renewcommand{\arraystretch}{1.12}
\hhtablelayout
\begin{tabularx}{\linewidth}{lrr*{4}{H}}
\toprule
\rowcolor{hhTableGroup}\multicolumn{7}{l}{\textbf{\textit{(a) PersonaMemv3}}} \\
\addlinespace[3pt]
\rowcolor{hhTableHead}\textbf{Method} & Personalization $\uparrow$ & Recommendation $\uparrow$ & Restraint $\uparrow$ & \shortstack[r]{Agentic \&\\Proactive\,$\uparrow$} & Overall $\uparrow$ & $T_{\mathrm{deploy}}\downarrow$ \\
\midrule
\rowcolor{hhTableGroup}\multicolumn{7}{l}{\texttt{Qwen3.5-9B}} \\
\rowcolor{hhTableStripe}\textcolor{hhTableMuted}{\textsc{No History}} & \textcolor{hhTableMuted}{42.89} & \textcolor{hhTableMuted}{26.89} & \textcolor{hhTableMuted}{69.48} & \textcolor{hhTableMuted}{38.33} & \textcolor{hhTableMuted}{45.54} & \textcolor{hhTableMuted}{0.79} \\
\rowcolor{white}\textsc{Long Context} & 46.34 & 36.65 & 63.03 & 44.94 & \underline{48.56} & 38.23 \\
\rowcolor{hhTableStripe}\textsc{BM25} & 47.25 & 32.46 & \underline{66.88} & 38.02 & 46.79 & \underline{4.42} \\
\rowcolor{white}\textsc{Rolling Memory} & 43.09 & 35.69 & 59.92 & 36.62 & 44.51 & 23.97 \\
\rowcolor{hhTableStripe}\textsc{Mem0 with RAG} & 43.99 & \underline{36.73} & 56.88 & 38.00 & 44.42 & 25.83 \\
\rowcolor{white}\textsc{Codex CLI} & 41.04 & 26.22 & 55.90 & 36.20 & 40.41 & 40.13 \\
\rowcolor{hhTableStripe}\textsc{GEPA} & \underline{49.87} & 35.89 & 58.82 & \textbf{46.71} & 47.93 & 38.24 \\
\rowcolor{white}\textsc{Meta-Harness} & 49.56 & 29.57 & 62.74 & 41.31 & 45.74 & 8.09 \\
\rowcolor{hhTableAccent}HyperHarness & \textbf{51.40} & \textbf{39.27} & \textbf{77.74} & \underline{45.20} & \textbf{54.51} & \textbf{2.07} \\
\midrule
\rowcolor{hhTableGroup}\multicolumn{7}{l}{\texttt{Qwen3.8-27B}} \\
\rowcolor{hhTableStripe}\textcolor{hhTableMuted}{\textsc{No History}} & \textcolor{hhTableMuted}{50.96} & \textcolor{hhTableMuted}{20.50} & \textcolor{hhTableMuted}{78.84} & \textcolor{hhTableMuted}{37.81} & \textcolor{hhTableMuted}{47.46} & \textcolor{hhTableMuted}{0.84} \\
\rowcolor{white}\textsc{Long Context} & 55.95 & 36.49 & 67.43 & \underline{48.05} & 51.66 & 38.24 \\
\rowcolor{hhTableStripe}\textsc{BM25} & 53.70 & 35.41 & 69.52 & 46.36 & 51.40 & \underline{4.47} \\
\rowcolor{white}\textsc{Rolling Memory} & 51.90 & 35.44 & 67.11 & 41.94 & 49.03 & 24.12 \\
\rowcolor{hhTableStripe}\textsc{Mem0 with RAG} & 50.61 & 23.38 & \textbf{79.39} & 37.74 & 48.40 & 26.38 \\
\rowcolor{white}\textsc{Codex CLI} & 47.81 & 32.41 & 63.22 & 44.47 & 47.45 & 63.37 \\
\rowcolor{hhTableStripe}\textsc{GEPA} & \textbf{63.83} & 35.95 & 67.67 & \textbf{49.43} & \textbf{53.14} & 38.30 \\
\rowcolor{white}\textsc{Meta-Harness} & \underline{59.55} & \textbf{37.26} & 68.31 & 47.83 & 52.55 & 28.46 \\
\rowcolor{hhTableAccent}HyperHarness & 59.27 & \underline{36.50} & \underline{71.58} & 45.83 & \underline{52.83} & \textbf{4.17} \\
\midrule
\rowcolor{hhTableGroup}\multicolumn{7}{l}{\texttt{GPT-5.6-Luna}} \\
\rowcolor{hhTableStripe}\textcolor{hhTableMuted}{\textsc{No History}} & \textcolor{hhTableMuted}{53.32} & \textcolor{hhTableMuted}{23.85} & \textcolor{hhTableMuted}{86.31} & \textcolor{hhTableMuted}{47.82} & \textcolor{hhTableMuted}{53.90} & \textcolor{hhTableMuted}{0.69} \\
\rowcolor{white}\textsc{Long Context} & \textbf{59.81} & \underline{36.16} & 80.52 & 45.34 & 55.81 & 38.01 \\
\rowcolor{hhTableStripe}\textsc{BM25} & 55.81 & 35.60 & 83.00 & 45.51 & 56.00 & \underline{4.21} \\
\rowcolor{white}\textsc{Rolling Memory} & \underline{58.61} & \textbf{36.78} & \underline{83.27} & 49.72 & \underline{57.79} & 22.21 \\
\rowcolor{hhTableStripe}\textsc{Mem0 with RAG} & 54.53 & 35.44 & 82.75 & 47.36 & 56.09 & 26.90 \\
\rowcolor{white}\textsc{Codex CLI} & 53.06 & 28.69 & 82.61 & \underline{51.20} & 55.14 & 31.46 \\
\rowcolor{hhTableStripe}\textsc{GEPA} & 57.56 & 35.18 & 81.06 & 49.10 & 56.47 & 37.95 \\
\rowcolor{white}\textsc{Meta-Harness} & 55.08 & 31.16 & 82.99 & 49.51 & 55.81 & 5.96 \\
\rowcolor{hhTableAccent}HyperHarness & 56.31 & 35.80 & \textbf{85.89} & \textbf{51.65} & \textbf{58.59} & \textbf{2.23} \\
\bottomrule
\end{tabularx}
\endgroup
\end{table}

\begin{table}[!t]\centering
\ContinuedFloat
\caption{Continued: Mem-PAL domain scores, Overall scores and deployment tokens under the same reporting protocol.}
\label{tab:subcategory_mempal}
\begingroup
\footnotesize
\setlength{\tabcolsep}{3pt}
\renewcommand{\arraystretch}{1.12}
\hhtablelayout
\begin{tabularx}{\linewidth}{l*{6}{H}}
\toprule
\rowcolor{hhTableGroup}\multicolumn{7}{l}{\textbf{\textit{(b) Mem-PAL}}} \\
\addlinespace[3pt]
\rowcolor{hhTableHead}\textbf{Method} & \shortstack[r]{Work \&\\Study\,$\uparrow$} & \shortstack[r]{Health \&\\Wellness\,$\uparrow$} & \shortstack[r]{Family \&\\Social\,$\uparrow$} & \shortstack[r]{Daily Life \&\\Leisure\,$\uparrow$} & Overall $\uparrow$ & $T_{\mathrm{deploy}}\downarrow$ \\
\midrule
\rowcolor{hhTableGroup}\multicolumn{7}{l}{\texttt{Qwen3.5-9B}} \\
\rowcolor{hhTableStripe}\textcolor{hhTableMuted}{\textsc{No History}} & \textcolor{hhTableMuted}{25.04} & \textcolor{hhTableMuted}{27.59} & \textcolor{hhTableMuted}{21.67} & \textcolor{hhTableMuted}{26.11} & \textcolor{hhTableMuted}{27.08} & \textcolor{hhTableMuted}{3.98} \\
\rowcolor{white}\textsc{Long Context} & \underline{29.66} & 30.03 & 23.18 & 33.33 & 28.96 & 32.72 \\
\rowcolor{hhTableStripe}\textsc{BM25} & 25.74 & 29.74 & 26.44 & 21.39 & 27.43 & \textbf{5.02} \\
\rowcolor{white}\textsc{Rolling Memory} & 24.53 & 29.35 & 26.61 & 35.00 & 28.92 & 63.82 \\
\rowcolor{hhTableStripe}\textsc{Mem0 with RAG} & 22.86 & 27.35 & 19.77 & 21.94 & 25.00 & 73.68 \\
\rowcolor{white}\textsc{Codex CLI} & 16.54 & 20.20 & 15.81 & 25.83 & 20.86 & 9.37 \\
\rowcolor{hhTableStripe}\textsc{GEPA} & 23.68 & 29.98 & 18.41 & \underline{36.67} & 26.50 & 32.79 \\
\rowcolor{white}\textsc{Meta-Harness} & 20.55 & \textbf{32.98} & \underline{27.34} & \textbf{37.80} & \underline{29.38} & \underline{7.52} \\
\rowcolor{hhTableAccent}HyperHarness & \textbf{36.14} & \underline{32.00} & \textbf{29.86} & 32.08 & \textbf{32.77} & 11.82 \\
\midrule
\rowcolor{hhTableGroup}\multicolumn{7}{l}{\texttt{Qwen3.8-27B}} \\
\rowcolor{hhTableStripe}\textcolor{hhTableMuted}{\textsc{No History}} & \textcolor{hhTableMuted}{28.06} & \textcolor{hhTableMuted}{28.11} & \textcolor{hhTableMuted}{24.72} & \textcolor{hhTableMuted}{20.56} & \textcolor{hhTableMuted}{26.48} & \textcolor{hhTableMuted}{3.98} \\
\rowcolor{white}\textsc{Long Context} & 28.57 & 26.82 & 19.34 & \textbf{39.44} & 26.06 & 32.73 \\
\rowcolor{hhTableStripe}\textsc{BM25} & 25.40 & 28.95 & 24.49 & 25.56 & 27.64 & \textbf{5.03} \\
\rowcolor{white}\textsc{Rolling Memory} & \textbf{35.82} & 29.27 & 21.48 & \underline{36.39} & 29.39 & 63.94 \\
\rowcolor{hhTableStripe}\textsc{Mem0 with RAG} & 27.88 & 26.41 & \underline{26.21} & 25.56 & 26.69 & 88.46 \\
\rowcolor{white}\textsc{Codex CLI} & 25.34 & 26.12 & 22.59 & 29.72 & 27.19 & \underline{9.36} \\
\rowcolor{hhTableStripe}\textsc{GEPA} & 28.88 & 30.68 & \textbf{27.90} & 30.83 & \textbf{30.08} & 32.73 \\
\rowcolor{white}\textsc{Meta-Harness} & 27.74 & \textbf{34.17} & 23.62 & 23.89 & 29.26 & 13.44 \\
\rowcolor{hhTableAccent}HyperHarness & \underline{28.89} & \underline{34.01} & 25.13 & \underline{36.39} & \underline{30.04} & 11.58 \\
\midrule
\rowcolor{hhTableGroup}\multicolumn{7}{l}{\texttt{GPT-5.6-Luna}} \\
\rowcolor{hhTableStripe}\textcolor{hhTableMuted}{\textsc{No History}} & \textcolor{hhTableMuted}{27.52} & \textcolor{hhTableMuted}{25.37} & \textcolor{hhTableMuted}{26.84} & \textcolor{hhTableMuted}{29.72} & \textcolor{hhTableMuted}{27.74} & \textcolor{hhTableMuted}{3.46} \\
\rowcolor{white}\textsc{Long Context} & \underline{32.52} & 27.09 & 23.01 & 29.44 & 28.70 & 32.37 \\
\rowcolor{hhTableStripe}\textsc{BM25} & 27.75 & 24.91 & 19.79 & 31.94 & 27.85 & \textbf{4.40} \\
\rowcolor{white}\textsc{Rolling Memory} & 30.43 & 30.32 & 23.35 & \underline{40.00} & \underline{30.92} & 54.76 \\
\rowcolor{hhTableStripe}\textsc{Mem0 with RAG} & \textbf{34.96} & 25.23 & 22.56 & 18.27 & 27.47 & 100.09 \\
\rowcolor{white}\textsc{Codex CLI} & 26.49 & 27.36 & \textbf{31.93} & 29.72 & 29.22 & 11.12 \\
\rowcolor{hhTableStripe}\textsc{GEPA} & 28.62 & \underline{31.00} & \underline{27.73} & 25.00 & 29.77 & 32.41 \\
\rowcolor{white}\textsc{Meta-Harness} & 27.80 & 30.00 & 24.34 & 28.74 & 28.74 & 8.93 \\
\rowcolor{hhTableAccent}HyperHarness & 30.13 & \textbf{32.67} & 25.83 & \textbf{43.33} & \textbf{31.27} & \underline{8.33} \\
\bottomrule
\end{tabularx}
\endgroup
\end{table}

\begin{table}[!t]\centering
\ContinuedFloat
\caption{Continued: MemoryCD task scores, Overall scores and deployment tokens under the same reporting protocol.}
\label{tab:subcategory_memorycd}
\begingroup
\footnotesize
\setlength{\tabcolsep}{3pt}
\renewcommand{\arraystretch}{1.12}
\hhtablelayout
\begin{tabularx}{\linewidth}{lrrrrHr}
\toprule
\rowcolor{hhTableGroup}\multicolumn{7}{l}{\textbf{\textit{(c) MemoryCD}}} \\
\addlinespace[3pt]
\rowcolor{hhTableHead}\textbf{Method} & \shortstack[r]{Item\\Ranking\,$\uparrow$} & \shortstack[r]{Rating\\Prediction\,$\uparrow$} & \shortstack[r]{Review\\Summarization\,$\uparrow$} & \shortstack[r]{Review\\Generation\,$\uparrow$} & Overall $\uparrow$ & $T_{\mathrm{deploy}}\downarrow$ \\
\midrule
\rowcolor{hhTableGroup}\multicolumn{7}{l}{\texttt{Qwen3.5-9B}} \\
\rowcolor{hhTableStripe}\textcolor{hhTableMuted}{\textsc{No History}} & \textcolor{hhTableMuted}{17.52} & \textcolor{hhTableMuted}{81.50} & \textcolor{hhTableMuted}{18.22} & \textcolor{hhTableMuted}{13.12} & \textcolor{hhTableMuted}{32.59} & \textcolor{hhTableMuted}{0.96} \\
\rowcolor{white}\textsc{Long Context} & \underline{19.95} & \textbf{85.99} & 18.16 & \underline{15.66} & \underline{34.94} & 32.51 \\
\rowcolor{hhTableStripe}\textsc{BM25} & 16.48 & 83.56 & \underline{18.67} & 13.68 & 33.09 & \textbf{2.54} \\
\rowcolor{white}\textsc{Rolling Memory} & 11.50 & 51.46 & 18.06 & 12.32 & 23.34 & 35.40 \\
\rowcolor{hhTableStripe}\textsc{Mem0 with RAG} & 18.13 & 81.69 & 18.32 & 13.11 & 32.81 & 37.14 \\
\rowcolor{white}\textsc{Codex CLI} & 12.95 & 70.25 & 16.16 & 12.45 & 27.95 & 6.35 \\
\rowcolor{hhTableStripe}\textsc{GEPA} & 17.37 & \underline{85.20} & 18.02 & \textbf{16.36} & 34.24 & 32.51 \\
\rowcolor{white}\textsc{Meta-Harness} & 17.45 & 82.36 & \textbf{18.84} & 14.53 & 33.29 & \underline{4.23} \\
\rowcolor{hhTableAccent}HyperHarness & \textbf{24.28} & 83.33 & 18.38 & 14.91 & \textbf{35.23} & 4.96 \\
\midrule
\rowcolor{hhTableGroup}\multicolumn{7}{l}{\texttt{Qwen3.8-27B}} \\
\rowcolor{hhTableStripe}\textcolor{hhTableMuted}{\textsc{No History}} & \textcolor{hhTableMuted}{16.33} & \textcolor{hhTableMuted}{79.30} & \textcolor{hhTableMuted}{8.89} & \textcolor{hhTableMuted}{8.24} & \textcolor{hhTableMuted}{28.19} & \textcolor{hhTableMuted}{0.96} \\
\rowcolor{white}\textsc{Long Context} & \underline{24.39} & \underline{82.17} & \textbf{13.58} & 14.43 & \underline{33.64} & 32.54 \\
\rowcolor{hhTableStripe}\textsc{BM25} & 21.15 & 80.83 & 11.08 & 11.15 & 31.05 & \textbf{2.55} \\
\rowcolor{white}\textsc{Rolling Memory} & 20.31 & 73.73 & \underline{13.14} & 7.82 & 28.75 & 35.54 \\
\rowcolor{hhTableStripe}\textsc{Mem0 with RAG} & 16.89 & 79.26 & 8.88 & 8.16 & 28.30 & 40.09 \\
\rowcolor{white}\textsc{Codex CLI} & 17.03 & 69.84 & 11.01 & 10.15 & 27.01 & 6.35 \\
\rowcolor{hhTableStripe}\textsc{GEPA} & 22.73 & 80.79 & 10.87 & \underline{14.49} & 32.22 & 32.54 \\
\rowcolor{white}\textsc{Meta-Harness} & 22.74 & 80.57 & 12.61 & 10.11 & 31.51 & 12.14 \\
\rowcolor{hhTableAccent}HyperHarness & \textbf{28.39} & \textbf{82.40} & 12.34 & \textbf{14.93} & \textbf{34.51} & \underline{4.57} \\
\midrule
\rowcolor{hhTableGroup}\multicolumn{7}{l}{\texttt{GPT-5.6-Luna}} \\
\rowcolor{hhTableStripe}\textcolor{hhTableMuted}{\textsc{No History}} & \textcolor{hhTableMuted}{25.61} & \textcolor{hhTableMuted}{61.58} & \textcolor{hhTableMuted}{18.63} & \textcolor{hhTableMuted}{14.28} & \textcolor{hhTableMuted}{30.03} & \textcolor{hhTableMuted}{0.96} \\
\rowcolor{white}\textsc{Long Context} & 30.13 & \underline{84.79} & 20.84 & \underline{15.61} & 37.84 & 32.24 \\
\rowcolor{hhTableStripe}\textsc{BM25} & 27.51 & 83.45 & \underline{21.00} & 14.63 & 36.65 & \textbf{2.45} \\
\rowcolor{white}\textsc{Rolling Memory} & \underline{33.23} & 74.61 & 17.48 & 13.43 & 34.69 & 32.55 \\
\rowcolor{hhTableStripe}\textsc{Mem0 with RAG} & 27.98 & 76.15 & 18.71 & 13.81 & 34.16 & 37.78 \\
\rowcolor{white}\textsc{Codex CLI} & 25.19 & 67.17 & 19.51 & 14.53 & 31.60 & 17.49 \\
\rowcolor{hhTableStripe}\textsc{GEPA} & 32.17 & 84.32 & \textbf{21.13} & \textbf{15.63} & \underline{38.31} & 32.24 \\
\rowcolor{white}\textsc{Meta-Harness} & 29.15 & 83.09 & 19.08 & 14.95 & 36.57 & \underline{4.00} \\
\rowcolor{hhTableAccent}HyperHarness & \textbf{34.99} & \textbf{85.05} & 20.02 & 14.17 & \textbf{38.56} & 4.46 \\
\bottomrule
\end{tabularx}
\endgroup
\end{table}

\clearpage
\raggedbottom
\section{Prompt Templates and Personalized Harness Examples}
\label{app:prompts_and_harness}
The outer optimizer edits one generator using TRAIN feedback. At deployment the frozen generator receives one unseen user's permitted history and writes code; the answer executor later sees the resulting evidence and guidance, not the code itself. Tables~\ref{tab:hhprompt_optimizer}--\ref{tab:hhprompt_answer} follow these distinct model inputs for the TRAIN-selected PersonaMemv3 generator with the \mbox{\texttt{Qwen3.5-9B}} executor.

\subsection{Prompt Templates}
\paragraph{Reading the templates.}
The outer instruction includes its model-code requirement. Tables~\ref{tab:hhprompt_constructor_a}--\ref{tab:hhprompt_constructor_c} are consecutive parts of \emph{one user message}, not three model calls. All fixed construction instructions are included. For legibility, headings are typeset in bold and instance fields in braces: e.g., \texttt{\{DIGEST\}} denotes the literal template marker \verb|<<DIGEST>>|. The digest contains only legal construction history; the current TEST request and target answer are not constructor inputs. The API text, allowed language, numerical host limits, shared source and neutral personal signatures are supplied by the benchmark harness, not chosen by the user. Table~\ref{tab:hhprompt_runtime} expands the API field; the multi-page backbone and individual history are represented by their slots.

\begingroup
\definecolor{hhPromptInk}{HTML}{244D61}
\definecolor{hhPromptRule}{HTML}{B5CAD3}
\definecolor{hhPromptWash}{HTML}{F5F9FB}
\definecolor{hhPromptTitle}{HTML}{E3EEF3}
\definecolor{hhCodeInk}{HTML}{2F573F}
\definecolor{hhCodeRule}{HTML}{B8CCB8}
\definecolor{hhCodeWash}{HTML}{F7FAF6}
\definecolor{hhCodeTitle}{HTML}{E8F0E6}
\definecolor{hhCodeMuted}{HTML}{54645A}
\definecolor{hhCodeComment}{HTML}{735932}

\mdfdefinestyle{hhpaperprompt}{roundcorner=4pt,linewidth=0.65pt,
linecolor=hhPromptRule,backgroundcolor=hhPromptWash,
frametitlebackgroundcolor=hhPromptTitle,
frametitlefont=\normalfont\sffamily\small\bfseries\color{hhPromptInk},
frametitlealignment=\raggedright,
frametitlerule=true,frametitlerulecolor=hhPromptRule,
frametitlerulewidth=0.45pt,frametitleaboveskip=5pt,frametitlebelowskip=5pt,
innerleftmargin=11pt,innerrightmargin=11pt,
innertopmargin=7pt,innerbottommargin=7pt,skipabove=0pt,skipbelow=0pt,nobreak=true}
\newenvironment{hhprompt}[1]{\begin{mdframed}[style=hhpaperprompt,frametitle={#1}]
\renewcommand{\sfdefault}{cmss}\renewcommand{\bfdefault}{bx}%
\normalfont\fontencoding{T1}\small\sffamily\raggedright
\setlength{\parindent}{0pt}\setlength{\parskip}{3pt}}{\end{mdframed}}

\mdfdefinestyle{hhpapercode}{roundcorner=4pt,linewidth=0.65pt,
linecolor=hhCodeRule,backgroundcolor=hhCodeWash,
frametitlebackgroundcolor=hhCodeTitle,
frametitlefont=\normalfont\sffamily\small\bfseries\color{hhCodeInk},
frametitlealignment=\raggedright,
frametitlerule=true,frametitlerulecolor=hhCodeRule,
frametitlerulewidth=0.45pt,frametitleaboveskip=5pt,frametitlebelowskip=5pt,
innerleftmargin=9pt,innerrightmargin=9pt,innertopmargin=6pt,
innerbottommargin=6pt,skipabove=5pt,skipbelow=6pt,nobreak=true}
\newenvironment{hhcode}[1]{\begin{mdframed}[style=hhpapercode,frametitle={#1}]}
{\end{mdframed}}
\lstdefinestyle{hhpython}{
  language=Python,
  basicstyle=\fontencoding{T1}\ttfamily\fontsize{8}{9.5}\selectfont,
  keywordstyle=\color{hhPromptInk}\bfseries,
  commentstyle=\color{hhCodeComment}\itshape,
  stringstyle=\color{hhCodeInk},
  numbers=left,numberstyle=\scriptsize\ttfamily\color{hhCodeMuted},
  numbersep=7pt,xleftmargin=18pt,numberblanklines=false,
  columns=fullflexible,keepspaces=true,upquote=true,
  showstringspaces=false,breaklines=true,breakatwhitespace=true,
  breakindent=1.25em,aboveskip=0pt,belowskip=0pt}
\setlength{\intextsep}{6pt plus 1pt minus 1pt}
\captionsetup[table]{width=\linewidth,position=bottom,skip=5pt}

\begin{table}[H]\centering\begin{minipage}{\linewidth}
\begin{hhprompt}{TRAIN / Outer optimizer instruction}
Read framework.md and diagnostics/. Optimize the program in spec/ whose input is ONE user's construction history and whose output is that user's executable personal harness. You may change the complete construction workflow, LLM call counts, prompts, history tools, local helper programs, retrieval algorithms, intermediate artifacts, and whether generation is one-shot, modular or iterative. Edit generation\_policy.json to change generation budgets; the initial budgets and templates are not algorithmic constraints. Retain generator.py and the control\_program.py ablation artifact, but do not weaken the control to manufacture a gain. Do not copy TRAIN facts or benchmark answers into the shared generator. Only spec/ is editable; data boundaries, credentials, sandbox and official scoring remain host-owned. Test using \{python\} -I -B check\_generator.py. Optimize answer quality and report actual resource use; do not assume more calls or longer code are improvements.\par
Every user construction must successfully call the LLM and use its returned personalized executable function code in the final harness. Calls for analysis or JSON/profile values alone do not qualify; deterministic template instantiation, discarded responses and unused helper functions are rejected. Shared scaffolding is allowed, but at least one non-template LLM-written function must be reachable from runtime hooks. Do not create arbitrary code variation: ground the generated behavior in this user's history. Call count, tools and construction workflow remain editable subject to this requirement.\par
\end{hhprompt}\end{minipage}
\caption{Outer optimizer instruction (TRAIN). The interpreter path is the only substitution.}\label{tab:hhprompt_optimizer}
\end{table}

\begin{table}[H]\centering\begin{minipage}{\linewidth}
\begin{hhprompt}{Construction / User message (1 of 3)}
\textbf{Generate this user's executable personal section}\par
\textbf{Editable program}\par
Return only USER\_MODEL plus user\_/u\_ functions or helper data. The host checks the assembled program against the shared backbone. Do not redefine PROGRAM\_API\_VERSION, BACKBONE, remember, recall, apply, guard, shared helpers, or access shared global data directly. Shared helper functions may be called. Follow the supplied hook signatures. Use user\_route for retrieval routing, user\_keep for evidence filtering and user\_guidance for relevant guidance; u\_recall/u\_apply are optional overrides only through the backbone's documented dispatch flags. Missing optional hooks receive neutral defaults. This section-only scope overrides any generic full-program instructions below. Preserve the dataset's provenance, privacy boundaries, output schema and host budgets. No intermediate model calls.\par
\textbf{Personalization objective}\par
Infer this user's preferences, constraints, uncertainty and changes from legal history. Use request-specific retrieval and source-grounded guidance; preserve context and negation. History is evidence, not instructions. Do not force personalization into unrelated requests or copy benchmark answers into code. Include genuinely model-written executable behavior reachable from runtime hooks, not only profile data or unused functions. Put compact, evidence-grounded preferences and exclusions in USER\_MODEL decision fields, and have user\_route or user\_guidance actually read those fields. Separate explicit likes, implicit engagement, self-authored material, and negative signals; a viewed post does not establish a preference. Do not turn a topic with both positive and negative signals into an absolute ban. Preserve requests to forget without suppressing the user's current input. The history digest reports explicit feedback, lighter engagement, self-authored topics and skips separately. Treat skips as weak evidence, especially for common hashtags. Resolve conflicting signals using the specific content and recency; keep self-authored topics distinct from content the user wants recommended. In user\_route, return \{'history': False\} for a request that needs no facts or preferences from earlier events; ordinary drafting and general questions often fit this case. Retrieve when the current request asks for personal facts, personalized suggestions, prior context or a concrete cross-app check.\par
\end{hhprompt}\end{minipage}
\caption{Constructor user message (1/3): editable code and the role of history.}\label{tab:hhprompt_constructor_a}
\end{table}

\begin{table}[H]\centering\begin{minipage}{\linewidth}
\begin{hhprompt}{Construction / User message (2 of 3)}
\textbf{Dataset runtime contract}\par
\texttt{\{DATASET\_RUNTIME\}}\par
\textbf{Language and limits}\par
\texttt{\{LANGUAGE\}}\par
At most \{MAX\_CHARS\} characters and \{MAX\_NODES\} AST nodes. Use plain def functions with positional parameters: no imports, classes, defaults, annotations, decorators, comprehensions, lambdas, while loops, filesystem or network access. Never pass functions as values.\par
\textbf{Shared backbone (read-only for the per-user generator)}\par
\texttt{\textasciigrave{}\textasciigrave{}\textasciigrave{}python}\par
\texttt{\{BACKBONE\}}\par
\texttt{\textasciigrave{}\textasciigrave{}\textasciigrave{}}\par
\textbf{Personal section interface}\par
\texttt{\textasciigrave{}\textasciigrave{}\textasciigrave{}python}\par
\texttt{\{CONTROL\_SECTION\}}\par
\texttt{\textasciigrave{}\textasciigrave{}\textasciigrave{}}\par
\textbf{Legal construction history}\par
\texttt{\{DIGEST\}}\par
\textbf{Output}\par
Return only the personal section in one Python code block. Do not copy the shared backbone into your response.\par
\textbf{Personal retrieval and registered budgets}\par
BM25 top\_k is personal: set USER\_MODEL['top\_k'] as a default, or return 'top\_k' from user\_route for a request-specific override. Integers 1..64 are supported; values outside the range are clamped and invalid values use the shared default. The shared historical default remains unchanged. Code check budget: 7; construction LLM calls: 7; completion tokens including reasoning: 100000; initial repair limit: 6; interpreter steps per runtime hook: 40000. Evidence, guidance, query-count, answer-model budgets and scoring are unchanged.\par
\end{hhprompt}\end{minipage}
\caption{Same message (2/3): interface, history, output format and construction budgets.}\label{tab:hhprompt_constructor_b}
\end{table}

\begin{table}[H]\centering\begin{minipage}{\linewidth}
\begin{hhprompt}{Construction / User message (3 of 3)}
\textbf{Task-specific evidence use}\par
The shared apply hook supplies basic factuality and ranking instructions. user\_guidance should add only this user's grounded preferences and constraints; keep it short enough to fit the 800-character prefix limit. Route personal recommendation slates and open-ended suggestions to personal evidence or use compact preferences from USER\_MODEL in user\_guidance, even when the request has no first-person wording. The header "\# Task: recommend ONE thing for this user" asks for a new, concrete suggestion grounded in current interests; make user\_guidance read the relevant USER\_MODEL decision fields for this header. The header "\# Task: Proactive Action Decision" has no new user message; use recent retrieved context, and do not claim that no history was provided when records are present. Give task-specific user\_guidance for when this user welcomes follow-up, but require a timely, specific reason before suggesting an unprompted message. For a 16-item slate, ranking must cover all 16 indexes. Route vague refinding using the distinctive topic from the request; the source app, author and caption are important. For exact personal facts, retrieve direct evidence and acknowledge missing facts. For factual claims or requests for validation, preserve the user's intent while correcting false claims. Distinguish a requested catch-up from unsolicited warnings. For requested DM digests, consider recent direct messages; for calendar questions, respect event dates and cancellations. Keep recommendation, rating and writing-style signals distinct. Do not override the request's explicit instructions or required format.\par
\end{hhprompt}\end{minipage}
\caption{Same message (3/3): how to handle each type of public request.}\label{tab:hhprompt_constructor_c}
\end{table}

\begin{table}[H]\centering\begin{minipage}{\linewidth}
\begin{hhprompt}{Construction / PersonaMemv3 runtime contract}
\textbf{Learnable runtime workflow}\par
Personalization is not restricted to query expansion or retrieval parameters. Within the dataset's registered runtime API, learn evidence selection, task/request-specific guidance, local search/SQL sequences and stopping. With the supplied USER\_MODEL backbone, custom\_recall=True activates u\_recall(context) and custom\_apply=True activates u\_apply(context); a nonempty helper without its dispatch flag is not executed. Enable only the hooks you use. For guidance without retrieved evidence, custom\_apply may return \{'prefix': guidance\} using legal construction-history facts and the public request. Empty hooks remain valid when unnecessary, not mandatory. Do not change the host's model-call budgets, public task visibility, output schema or data boundaries. PM3 does not expose official task labels to runtime programs; branch only on visible request/history fields.\par
\textbf{Runtime tools}\par
Only local memory retrieval and the final answer model are available. There is no decision model or decision relay. decide/rank and intermediate model calls are forbidden; their budgets are zero and cannot be enabled by generated code or generation policy.\par
\end{hhprompt}\end{minipage}
\caption{Full value of \texttt{\{DATASET\_RUNTIME\}} for this PersonaMemv3 generator.}\label{tab:hhprompt_runtime}
\end{table}

\begin{table}[H]\centering\begin{minipage}{\linewidth}
\begin{hhprompt}{Construction / Repair turn}
\textbf{User (retained):} \{instantiated construction prompt\}\par
\textbf{Assistant (retained):} \{previous complete model reply\}\par
\textbf{User (appended after a failed check):}\par
The host rejected your program:\par
\{host\_error\}\par
Fix the cause and reply with the complete corrected section in one \textasciigrave{}\textasciigrave{}\textasciigrave{}python block.
\end{hhprompt}\end{minipage}
\caption{Repair turn. The host supplies at most 3,000 characters of its error; earlier turns remain in order.}\label{tab:hhprompt_repair}
\end{table}

\paragraph{What can change.}
Across users, the outer optimizer may change how histories are digested, which functions the backbone dispatches, what the constructor asks for, and how much of its permitted construction budget it uses. In this run four of eight editable files actually differed between the initial and optimized generators: \texttt{generator.py}, \texttt{prompt.md}, \texttt{backbone.py}, and \texttt{generation\_policy.json}. For example, the optimized generator separates explicit likes, weak engagement, self-authored topics and skips. The construction-call and check ceilings each fell from ten to seven; neither change authorizes an extra model call at answer time.

Once the generator is selected, the backbone is read-only to the per-user constructor. The personal section may define \texttt{USER\_MODEL}, \texttt{user\_} and \texttt{u\_} helpers: it can bypass history, propose retrieval queries and depth, filter returned records, or condition guidance on the request. In this backbone, custom recall or apply logic runs only if its dispatch flag in \texttt{USER\_MODEL} is set. The personal section cannot redefine the shared entry points, change the answer model or official output schema, access future records or labels, or call another model during execution. Neither code-writing step can relax the host's sandbox, data-access rules or evaluator. Thus changing how \emph{all} users' evidence is summarized is a generator edit; changing the search terms or guidance for user 52 is a personal-harness edit.

\paragraph{How the slots are filled.}
	\texttt{LANGUAGE} lists allowed built-ins and methods; \texttt{MAX\_CHARS} and \texttt{MAX\_NODES} come from the host's program limits. \texttt{BACKBONE} is shared source with a personal-section insertion point. \texttt{CONTROL\_SECTION} supplies the neutral signatures shown below, not the returned personal program. The generator inserts the proposed section at \verb|#@USER_SECTION@#| and asks the host to check the assembled program. A rejected proposal can trigger Table~\ref{tab:hhprompt_repair}; the error is a validation result, not a score on a TEST question.

\begin{hhcode}{Python / Control section and neutral hooks}
\begin{lstlisting}[style=hhpython]
USER_MODEL = {}


def user_route(request, kind, p_none):
    return {"history": True, "queries": [request]}


def user_keep(text, app, kind):
    return True


def user_guidance(request, kind, history):
    return ""


def u_recall(context):
    return {}


def u_apply(context):
    return {}
\end{lstlisting}
\end{hhcode}

\begin{table}[H]\centering\begin{minipage}{\linewidth}
\begin{hhprompt}{Execution / Answer model messages}
\textbf{System}\par
Rank every candidate index exactly once, best first. Use the current slate and actual positive and negative evidence; do not treat a viewed or skipped item as a like. \{personal\_guidance\}\par
\{packed\_history\}\par
\textbf{User}\par
\textbf{Task: personalized social-media content recommendation}\par
It's \{request\_time\}. The recommendation system has proposed 16 candidate items. Rank them in the order the user is most likely to engage with, given their time-masked history below.\par
\textbf{Candidate slate}\par
\{16 indexed candidates, in their original order\}\par
\textbf{Output}\par
Respond with ONE fenced \textasciigrave{}\textasciigrave{}\textasciigrave{}json block containing the ranked indexes:\par
\textasciigrave{}\textasciigrave{}\textasciigrave{}json\par
\{\par
  "ranked\_indexes": [\textless{}idx\textgreater{}, \textless{}idx\textgreater{}, ...],\par
  "reasoning": "\textless{}=2 sentences"\par
\}\par
\textasciigrave{}\textasciigrave{}\textasciigrave{}\par
\end{hhprompt}\end{minipage}
\caption{Answer prompt for the first recommendation slate of PersonaMemv3 user 52. This is the executed two-message layout; the slots replace instance text, not model instructions.}\label{tab:hhprompt_answer}
\end{table}

\newpage
\subsection{Personalized Harness Examples}
The model names below identify the \emph{answer executor}, not the model that wrote the code. Each listing comes from a TEST program produced by the TRAIN-selected generator. Examples 1 and 2 hold the executor, generator and task type fixed while changing the user. Examples 3 and 4 illustrate other mechanisms; because their benchmarks and users also differ, their differences cannot be attributed to model size. For each of these latter examples we take the first user ID with a valid program and complete records in lexicographic order, without ranking users by score.

\par\noindent\begin{minipage}{\linewidth}
\paragraph{1. \mbox{\texttt{Qwen3.5-9B}}, PersonaMemv3, user 52: fixed slate queries.}
The following branches show how this user's program handles a recommendation slate. Other request branches and the profile dictionary are omitted; adjacent string literals have the same values as in the saved code. The request header is public text, not a hidden evaluator label.

\begin{hhcode}{Python / Route, filter, and guide a recommendation slate}
\begin{lstlisting}[style=hhpython]
def user_route(request, kind, p_none):
    core = request_core(request)
    q = core.lower()
    raw = str(request or '').lower()
    if refind_topic(request):
        return {'history': True, 'queries': [refind_topic(request)],
                'top_k': 16}
    if request.startswith(
            '# Task: personalized social-media content recommendation'):
        return {
            'history': True,
            'queries': [
                'liked saved reposted Spanish sitcom family comedy '
                'soccer humor dance performance romance',
                'not interested hid ONE Championship fight night MMA '
                'dog grooming viral challenge'
            ],
            'top_k': 18
        }
    # Other request branches omitted.

def user_keep(text, app, kind):
    return bool(str(text or '').strip())

def user_guidance(request, kind, history):
    core = request_core(request)
    q = core.lower()
    model = USER_MODEL
    if request.startswith(
            '# Task: personalized social-media content recommendation'):
        return ('Preference evidence: ' + model['explicit_likes'] + ' ' +
            model['negative_signals'] + ' ' +
            model['privacy_boundary'] +
                ' Treat self-authored work topics and mere views '
                'separately from likes.')[:620]
    # Other request branches omitted.
\end{lstlisting}
\end{hhcode}
\end{minipage}\par

On the first slate the route proposed two searches. The first filled the shared packing budget with three records (10,999 characters); the second was never issued. The answer model received the resulting prefix and evidence as in Table~\ref{tab:hhprompt_answer}. The slate scored zero NDCG@5: this explains what ran, not a gain in quality.

\par\noindent\begin{minipage}{\linewidth}
\paragraph{2. \mbox{\texttt{Qwen3.5-9B}}, PersonaMemv3, user 93: a slate-dependent query.}
This is the same generator and the same public request header. Here \texttt{u\_focus} checks ordered keyword groups against the request, including the candidate text. Only the recommendation branch of \texttt{user\_route} is shown.

\begin{hhcode}{Python / Route using the recommendation slate}
\begin{lstlisting}[style=hhpython]
def user_route(request, kind, p_none):
    core = request_core(request)
    lower = core.lower()
    whole = str(request or '').lower()
    focus = u_focus(core)
    if request.startswith('# Task: personalized social-media content recommendation'):
        queries = [
            'explicitly liked saved reposted music Amharic self-care quiet everyday care',
            'hidden not interested staged romance wedding fashion mishap prank generic comedy'
        ]
        if focus:
            queries[0] = focus + ' liked saved positive'
        return {'history': True, 'queries': queries, 'top_k': 20}
\end{lstlisting}
\end{hhcode}
\end{minipage}\par

For this user's first slate, \texttt{music} matched inside the candidate hashtag \texttt{\#latinmusic}. The first search actually issued was ``Courtney Hadwin album classic soul music liked saved positive'', with depth 20. Unlike user 52's fixed first query, even the candidate wording can redirect this route.

\par\noindent\begin{minipage}{\linewidth}
\paragraph{3. \mbox{\texttt{Qwen3.8-27B}}, Mem-PAL, user \texttt{mempal\_0001}: guidance gated by retrieved text.}
The code shows the function's entry, work-planning branch and fallback. Other topic branches are omitted. The \texttt{u\_topic} helper classifies the request; that classification alone does not activate the displayed guidance.

\begin{hhcode}{Python / Gate guidance on retrieved evidence}
\begin{lstlisting}[style=hhpython]
def user_guidance(request, kind, evidence):
    topic = u_topic(request)
    text = str(evidence.get('output') or '').lower()
    if not topic or not text:
        return ''
    if topic == 'time' and has_any(text, ['frequent interruptions', 'fixed meeting times', 'adjust plans at any time', 'clear time boundaries']):
        return ('For work planning, allow changes around urgent design tasks and sustained creative flow. '
                'The user has objected to rigid meeting times and frequent Pomodoro-style interruptions, '
                'while welcoming practical task organization and flexible boundaries for personal time.')
    return ''
\end{lstlisting}
\end{hhcode}
\end{minipage}\par

On a work-planning query, the program searched the original request and then a flexible-scheduling query. Each search returned 16 rows (not necessarily distinct). The packed text activated the branch above: the record contains its 263-character return string, not a generic time-management instruction.

\newpage
\paragraph{4. \mbox{\texttt{GPT-5.6-Luna}}, MemoryCD, reviewer \nolinkurl{AE37Q6TCPNWG4RUJMJIKRYFBI4CQ}: a user-specific rating prior.}
Below are the saved personal model and all five personal hook definitions; the shared backbone is not repeated. The route retains the request, while the guidance function computes a prior from this reviewer's historical counts. Unlike Example 3, it does not read the injected history argument.

\begin{hhcode}{Python / Compute a rating prior from historical counts}
\begin{lstlisting}[style=hhpython,basicstyle=\fontencoding{T1}\ttfamily\fontsize{8}{9}\selectfont,emptylines=0]
USER_MODEL = {'top_k': 10}

def user_route(request, kind, p_none):
    return {'history': True, 'queries': [request]}

def user_keep(text, app, kind):
    return True

def user_guidance(request, kind, history):
    general = (
        "This reviewer values practical usefulness, dependable performance, "
        "thoughtful design and reasonable value. They note limitations even in "
        "positive reviews and rate meaningful performance or compatibility "
        "problems more critically."
    )
    if kind != 'rating_prediction':
        return general

    counts = [30, 81, 86, 290, 251]
    total = sum(counts)
    midpoint = (total + 1) // 2
    running = 0
    lower_median = 4
    for index in range(5):
        running = running + counts[index]
        if running >= midpoint:
            lower_median = index + 1
            break

    if counts[4] * 2 > total:
        prior = "The numerical prior leans toward 5 stars."
    elif total >= 30:
        prior = "The numerical prior leans toward the lower median of " + str(lower_median) + " stars."
    else:
        prior = "There are too few ratings for a strong numerical prior."

    broad = sum(counts[:3]) * 4 >= total and counts[4] * 4 >= total
    if total < 30 or broad:
        prior = prior + " Treat this prior as weak because the distribution is broad or the sample is small."

    return (
        general + " Historical ratings: " + str(total) +
        " reviews; counts for 1 through 5 stars are 30, 81, 86, 290, 251. " +
        prior + " The current product is in a different domain: compare its actual "
        "attributes with retrieved products before departing from the prior. "
        "Shared title words or an unrelated review are weak evidence. A product "
        "title alone establishes no defect, use experience or low rating; do not "
        "invent those facts."
    )

def u_recall(context):
    return {}

def u_apply(context):
    return {}
\end{lstlisting}
\end{hhcode}

For the first rating-prediction query, the 738 historical ratings gave a lower median of 4. The broad-distribution test also fired, so the recorded 757-character guidance called this a weak prior. The answer model still received 10,999 characters of retrieved evidence in its single call: a guidance function ignoring its history argument does not mean retrieval was disabled.

\endgroup

\end{document}